\documentclass[letterpaper,11pt]{article}
\usepackage{newtxtext}
\usepackage[libertine,cmbraces]{newtxmath} 
\let\forall\forallAlt  

\usepackage[left=2.8cm,right=2.8cm]{geometry}

\providecommand{\keywords}[1]
{
  \small	
  \textbf{\textit{Key words : }} #1
}

\usepackage{endnotes}
\let\footnote=\endnote

\usepackage[utf8]{inputenc}

\usepackage{amsmath}
\usepackage{caption}
\usepackage{enumerate}
\usepackage{subcaption}
\usepackage[edges]{forest}
\usepackage{xcolor}
\usepackage{color, colortbl}
\usepackage{blkarray}
\usepackage{float}
\usepackage{tikz}

\newcommand{\norm}[1]{\ensuremath{\bar{#1}}}		

\sloppypar

\date{}

\usepackage{natbib}
 \bibpunct[, ]{(}{)}{,}{a}{}{,}%

\title{Explaining Results of Multi-Criteria Decision Making}

\author{Martin Erwig, Prashant Kumar\\
\{erwig,kumarpra\}@oregonstate.edu
}
\begin{document}

\maketitle

\begin{abstract}

We introduce a method for explaining the results of various linear and hierarchical multi-criteria decision-making (MCDM) techniques such as WSM and AHP. 
The two key ideas are (A) to maintain a fine-grained representation of the values manipulated by these techniques and (B) to derive explanations from these representations through merging, filtering, and aggregating operations.
An explanation in our model presents a high-level comparison of two alternatives in an MCDM problem, presumably an optimal and a non-optimal one, illuminating why one alternative was preferred over the other one. 
We show the usefulness of our techniques by generating explanations for two well-known examples from the MCDM literature.
Finally, we show their efficacy by performing computational experiments.

\end{abstract}








\maketitle

%


%
%
%


\captionsetup[figure]{font=small}
\captionsetup[subfigure]{font=small}

\section{Introduction}
\label{IntroAHP}

The theory and methods of multi-criteria decision-making (MCDM) have been extensively applied in many areas, ranging from engineering projects, economics, public administration, to management and military projects.
For example, in 1986 the Institute of Strategic Studies in Pretoria, a government-backed organization, used the Analytic Hierarchy Process (AHP)~\citep{SAATY19909,SAATY1987161} to analyze the conflict in South Africa and recommended actions ranging from the release of Nelson Mandela to the removal of apartheid and the granting of full citizenship and equal rights to the black majority~\citep{articleSaatySA}. All the recommendations were implemented within a short time. Another high-profile example is the use of AHP in the 1995 US/China conflict over Chinese illegal copying of music, video, and software~\citep{SAATY2001243}. An AHP analysis involving four hierarchies for benefits, costs, opportunities, and risks showed, surprisingly, that it was much better for the United States \emph{not} to sanction China. The result of the study predicted what happened. Shortly after the study was complete, the United States awarded China the most-favored nation status and didn't sanction it.
In the domain of business, the Xerox Corporation has used AHP to allocate almost a billion dollars to its research projects~\citep{Saaty2002DECISIONMW}, and IBM used AHP in 1991 in designing its successful mid-range AS 400 computer~\citep{TANG199222}.

Given the wide-spread use and impact of MCDM methods, it is natural for decision makers to ask why a certain alternative was suggested. Automation systems can earn their users' trust by \emph{explaining} results because explanations give users confidence in the correctness and reliability of computation processes \citep{10.1145/3473856.3473886}, in particular when the recommended actions are counter-intuitive, such as the recommendation not to sanction China in the trade conflict. 
However, existing MCDM techniques do not provide explanations about their recommended actions. 

With the ever-increasing impact of algorithms and mathematical models in our daily lives, there has been an increased demand for their explainability, so much so that countries have started to incorporate the \emph{right to explanation}~\citep{10.1093/idpl/ipx022} of an algorithm's result impacting the life of its citizens as part of their legal framework. The General Data Protection Regulation of the European Union~\citep{GDPR2017} and the Digital Republic Act of France~\citep{FranceDigital} are examples of such regulations. This trend is expected to continue, and therefore it stands to reason that sooner rather than later there will be an expectation, if not an obligation, for modeling tools such as MCDM to explain their results. 

When presented with a result by an algorithm, a natural question for users to ask is ``Why is the result $X$ and not $Y$?''. Such a question calls for what the philosophy literature calls a \emph{contrastive explanation} \citep{Lipton04,Lipton90}. A contrastive explanation compares two specific phenomena, the actual result (called \emph{fact} or \emph{solution}) and a hypothetical alternative (called \emph{foil}) and justifies ``Why this [fact] rather than that [foil]?''~\citep{Garfinkel81}. 
A nice property of contrastive explanations is that they can be tailored to different users who may be wondering about different aspects of a solution. 
Different (parts of) results may result in different foils and consequently in different explanations. 
%

In this paper, we present a method to automatically generate explanations for the various forms of MCDM, using an explanation mechanism called \emph{minimal dominating sets (MDS)}~\citep{EK21jfp}, which allows explanations to focus on only the most relevant aspects of a decision, thereby allowing the generation of concise explanations. 
The main contributions of this paper are the following.

\begin{itemize}

\item An MDS-based \emph{explanation method} for the two linear MCDM techniques Weighted Sum Method (WSM)~\citep{FishburnWSM} and Weighted Product Method (WPM)~\citep{doi:10.1287/ited.2013.0124}. 

\item \emph{Hierarchical value decomposition} as a novel representation that facilitates the generalization of MDS explanations to the hierarchical MCDM technique AHP.

\item \emph{Experimental evidence} for the \emph{effectiveness} of our approach. 
\end{itemize}
The rest of this paper is structured as follows.
After reviewing the various MCDM techniques in Section \ref{lab:IntroMCDM}, we present in Section \ref{sec:theory} the MDS explanation technique for linear MCDM using linear value decomposition. 
In Section \ref{sec:theoryAHP}, we introduce hierarchical value decomposition as a generalization of the linear case. 
In Section \ref{sec:Hierarchical_explainataion}, we demonstrate 
how MDS explanations can be extended to work with this hierarchical structure. 
In Section \ref{sec:coarsegrained}, we introduce a method for simplifying explanations to a more coarse-grained form, thereby making them easier to understand. 
In Section \ref{sec:CaseStudy} we apply our explanation techniques to two examples from the MCDM literature, and in Section \ref{sec:eval} we evaluate the effectiveness of the MDS explanation mechanism. 
We discuss related work in Section \ref{sec:RelatedWork} and present conclusions in Section \ref{sec:Conclusions}.

\section{Multi-Criteria Decision Making}
\label{lab:IntroMCDM}

MCDM is the process of making decisions in the presence of multiple, usually conflicting, criteria. In this context, an \emph{alternative} represents one of several choices available to the decision maker. The goal of MCDM is to identify the best alternative. Each MCDM problem is associated with multiple \emph{attributes}, which represent the decision criteria. An attribute can be beneficial or detrimental to an alternative. Each attribute has an associated \emph{weight} that signifies the importance of that attribute. The weight of an attribute remains constant across all alternatives and varies between 0 and 1. All the attribute weights should sum to 1. 


When the number of attributes gets large, they can be arranged in a hierarchical manner so that attributes higher up in the hierarchy, also called \emph{major attributes}, aggregate the contributions of minor attributes that appear lower in the hierarchy. 
In Section \ref{sec:linearMCDM} we describe the two linear MCDM techniques WSM and WPM, followed by the hierarchical technique AHP in Section \ref{sec:hierarcicalMCDM}.

\newcommand{\treelft}{\langle}
\newcommand{\treergt}{\rangle}
\newcommand{\premisepat}[1]{\ensuremath{\cdots#1\cdots}}
\newcommand{\tree}[2]{\ensuremath{#1\treelft#2\treergt}}
\newcommand{\subtree}[2]{\tree{#1}{\premisepat{#2}}}

\newcommand{\categories}{\ensuremath{\mathbf{C}}}
\newcommand{\trees}{\ensuremath{\mathbf{T}}}

\newcommand{\totalSym}{\hat}
\newcommand{\total}[1]{\totalSym{#1}}

\newcommand{\valSym}{\ensuremath{\varphi}}
\newcommand{\val}[1]{\ensuremath{\valSym(#1)}}
\newcommand{\tval}[1]{\ensuremath{\totalSym{\valSym}(#1)}}

\newcommand{\pickSym}{\ensuremath{\succ}}
\newcommand{\pickRel}[1][\valSym]{\mathrel{\stackon[0pt]{\pickSym}{$\scriptscriptstyle#1$}}}
\newcommand{\pick}[3][\valSym]{\ensuremath{#2\pickRel[#1]#3}}

\newcommand{\leveled}[2][\ell]{\ensuremath{#2_{#1}}}

\newcommand{\diffSym}{\ensuremath{\delta}}
\newcommand{\diff}[2][{}]{\ensuremath{\leveled[#1]{\diffSym}(#2)}}
\newcommand{\xdiff}[2]{\ensuremath{\diffSym^#1}(#2)}
\newcommand{\tdiff}[2][{}]{\ensuremath{\leveled[#1]{\total{\diffSym}}(#2)}}
\newcommand{\barrSym}{\ensuremath{\beta}}
\newcommand{\barr}[2][{}]{\ensuremath{\leveled[#1]{\barrSym}(#2)}}
\newcommand{\lbarr}[2][{}]{\ensuremath{\leveled[#1]{\barrSym^\ell}(#2)}}
\newcommand{\xbarr}[2]{\ensuremath{\barrSym^#1}(#2)}
\newcommand{\barrl}[2]{\ensuremath{\barrSym_{#1}(#2)}}
\newcommand{\tbarr}[2][{}]{\ensuremath{\leveled[#1]{\total{\barrSym}}(#2)}}
\newcommand{\tbarrl}[2]{\ensuremath{\total{\barrSym_{#1}}(#2)}}

\newcommand{\resolSym}{\ensuremath{\Gamma}}
\newcommand{\resol}[2][{}]{\ensuremath{\leveled[#1]{\resolSym}(#2)}}
\newcommand{\lresol}[2][{}]{\ensuremath{\leveled[#1]{\resolSym^\ell}(#2)}}
\newcommand{\xresol}[2]{\ensuremath{\resolSym^#1}(#2)}
\newcommand{\resoll}[2]{\ensuremath{\resolSym_{#1}(#2)}}

\newcommand{\tresol}[1]{\ensuremath{\total{\resolSym}(#1)}}

\newcommand{\domiSym}{\ensuremath{\Delta}}
\newcommand{\domi}[2][{}]{\ensuremath{\leveled[#1]{\domiSym}(#2)}}
\newcommand{\ldomi}[2][{}]{\ensuremath{\leveled[#1]{\domiSym^\ell}(#2)}}
\newcommand{\xdomi}[2]{\ensuremath{\domiSym^#1}(#2)}
\newcommand{\domil}[2]{\ensuremath{\domiSym_{#1}(#2)}}

\newcommand{\minimal}[1]{\underline{#1}}
\newcommand{\mdomiSym}{\minimal{\domiSym}}
\newcommand{\mdomi}[2][{}]{\ensuremath{\leveled[#1]{\mdomiSym}(#2)}}
\newcommand{\lmdomi}[2][{}]{\ensuremath{\leveled[#1]{\mdomiSym^\ell}(#2)}}
\newcommand{\xmdomi}[2]{\ensuremath{\mdomiSym^#1}(#2)}
\newcommand{\mdomil}[2]{\ensuremath{\mdomiSym_{#1}(#2)}}

\newcommand{\catfmt}[1]{\textsc{\lowercase{#1}}}

\newcommand{\honda}{\catfmt{Honda}}
\newcommand{\bmw}{\catfmt{BMW}}
\newcommand{\friends}{\catfmt{Personal}}
\newcommand{\personal}{\catfmt{Personal}}
\newcommand{\experts}{\catfmt{Expert}}
\newcommand{\price}{\catfmt{P}}
\newcommand{\fuel}{\catfmt{FE}}
\newcommand{\safety}{\catfmt{SR}}
\newcommand{\comfort}{\catfmt{C}}
\newcommand{\priority}{\catfmt{T}}
\newcommand{\cardecomp}[4]{\ensuremath{%
  \{\price\mapsto#1,\fuel\mapsto#2,\safety\mapsto#3,\comfort\mapsto#4\}}}

\newcommand{\cardecompAHP}[3]{\ensuremath{%
  \{\price\mapsto#1,\fuel\mapsto#2,\safety\mapsto#3\}}}

\newcommand{\cardecompAHPL}[2]{\ensuremath{%
  \{\friends\mapsto#1,\experts\mapsto#2\}}}

\subsection{Linear MCDM}
\label{sec:linearMCDM}

A linear MCDM problem with $m$ alternatives and $n$ attributes can be expressed by an $m \times n$ matrix, called a \emph{decision matrix}. Row $A_j$ represents the attribute values of the $j^{\textrm{th}}$ alternative, and row $C_i$ represents the values of $i^{\textrm{th}}$ attribute for different alternatives, where $w_i$ is the weight of the $i^{\textrm{th}}$ attribute. Each value $a_{ji}$ is the value of the attribute $C_i$ for the alternative $A_j$. 
\[
\begin{blockarray}{ccccc}
& \mathbf{C_1} & \mathbf{C_2} & \ldots & \mathbf{C_n} \\
 & w_1 & w_2 &  & w_n \\
\begin{block}{c[cccc]}
  \mathbf{A_1} & a_{11} & a_{12} &        & a_{1n}  \\
  \mathbf{A_2} & a_{21} & a_{22} & \ldots & a_{2n}  \\
  \vdots &     &\vdots  & \ddots & \vdots   \\
 \mathbf{A_m} & a_{m1} & a_{m2} & \ldots & a_{mn}\\
\end{block}
\end{blockarray}
\]
As an example, consider the task of deciding which car to buy. Some attributes to be considered are \emph{price}, \emph{fuel efficiency}, \emph{safety rating}, and \emph{comfort}. A decision matrix for this MCDM problem is shown in Figure \ref{fig:wsm_Car}. Price is a detrimental attribute, which is indicated by a minus sign after the weight, whereas all other attributes are beneficial, as indicated by the plus sign. 

\newcommand{\smallbf}[1]{\small{\textbf{#1}}}
\begin{figure}[t]
\centering
\begin{small}
    \begingroup
    \setlength{\tabcolsep}{5pt} 
    %
    \begin{tabular}{r|cccc}
     &\smallbf{Price} & \smallbf{Efficiency} & \smallbf{Safety}  &  \smallbf{Comfort} \\
    \textbf{}  &  0.1 ($-$)  &  0.4 ($+$)  &  0.3 ($+$)  &  0.2 ($+$) \\[2pt]
    \hline  
    \smallbf{Toyota}  & \$22,000 & 32 & 8.5 & 6.7 \\[2pt] 
    \smallbf{Honda}   & \$25,000 & 38 & 7.5 & 7.9 \\[2pt]
    \smallbf{BMW}     &  \$27,000 & 35 & 9.6 & 9.2
    \end{tabular}
    \endgroup
\end{small}
\caption{WSM example: decision matrix for car selection}
\label{fig:wsm_Car}
\end{figure}

In general, the values for the different attributes are represented in different units and at different scales. For instance, cost is measured in dollars whereas fuel efficiency is measured in miles per gallon. 
To compare and combine values of different attribute values, all values have to be normalized as follows.
For a beneficial attribute $C_k$ and the alternative $A_l$, the normalized attribute value $\norm{a}_{lk}$ corresponding to $a_{lk}$ is defined as
$\norm{a}_{lk} = a_{lk}/\max_{1\leq j\leq m} a_{jk}$.
Similarly, in case the attribute $C_k$ is detrimental, the attribute value $\norm{a}_{lk}$  is defined as
$\norm{a}_{lk} = \min_{1\leq j\leq m} a_{jk}/a_{lk}$.





With a normalized decision matrix we can compute the optimal solution for a problem as follows.
%
%
The contribution of each attribute is obtained by a linear binary function of its weight and value. The particulars of the binary function depend on the specific MCDM method that is employed. The so-called \emph{score} $\tau$ of an alternative is the aggregation of the contributions of its attributes. For WSM, the contribution of attribute $C_i$ for alternative $A_j$ is given by $w_i \norm{a}_{ji}$. If each alternative consists of $n$ attributes, the score for $A_j$ in the WSM is given by:
\[
\tau_{A_j} = \sum_{1\leq i\leq n} w_i \norm{a}_{ji}
\]
For WPM the contribution of $C_i$ for alternative $A_j$ is given by $\norm{a}_{ji}^{w_i}$, and the score of $A_j$ is given by:
\[
\tau_{A_j} = \prod_{1\leq i\leq n} \norm{a}^{w_i}_{ji}
\]
The alternative with the highest score is by definition the best alternative. In Figure \ref{fig:ScoreCar} we present the normalized decision matrix with the aggregate scores in the last column for the car selection problem. The numbers suggest to by a BMW despite it being the most expensive car.

\begin{figure}[t]
\centering
\begin{small}
    \setlength{\tabcolsep}{5pt} 
    \begin{tabular}{r|cccc|c}
     &\smallbf{Price} & \smallbf{Efficiency} & \smallbf{Safety}  &  \smallbf{Comfort} & \smallbf{Score} \\
     &  0.1 ($-$)  &  0.4 ($+$)  &  0.3 ($+$)  &  0.2 ($+$)  &  $\tau$ \\[2pt]
    \hline  
    \smallbf{Toyota}  &  1    &	0.84 & 0.89 & 0.73 & 0.85 \\[2pt] 
    \smallbf{Honda}   &  0.88 &	1    & 0.78 & 0.86 & 0.89 \\[2pt]
    \smallbf{BMW}     &  0.81 & 0.92 & 1    & 1    & 0.95
    \end{tabular}
\end{small}
\caption{Normalized decision matrix for car selection with final score}
\label{fig:ScoreCar}
\end{figure}

\subsection{Hierarchical MCDM}
\label{sec:hierarcicalMCDM}

The idea behind the Analytic Hierarchy Process (AHP) is the decomposition of a complex problem into a hierarchy.
The leaves at the bottom of the hierarchy present the different alternatives, whereas internal nodes play a dual role: On one hand, they are attributes of the alternatives for the level immediately below them, on the other hand, they represent alternatives for the level above them.
AHP consists of the following three steps \citep{articleHarkerVargas,bookSattyVargas}.



\tikzstyle{startstop} = []
\tikzstyle{arrow} = []
\begin{figure}[b]
    \centering
        \begin{tikzpicture}[node distance=1.2cm]
        level 1/.style={level distance=0.5cm}
        \node (total) [startstop] {Score};
        \node (friends) [startstop, below of = total, xshift = -1.2cm] {Personal} ;
        \node (experts) [startstop, right of = friends,xshift = 1.2cm] {Expert}; 
        \node (price) [startstop, below of = friends,  xshift = -1.5cm] {Price};
        \node (fuel) [startstop, right of = price,  xshift = 1.0cm] {Fuel Efficiency}; 
        \node (safety) [startstop, right of = fuel,xshift = 1.9cm] {Safety Ratings};
        \node (honda) [startstop, below of = fuel,  xshift = -0.1cm] {Honda}; 
        \node (bmw) [startstop, right of = honda,xshift = 1.0cm] {BMW};
        \draw [arrow] (total.south) -- (friends);
        \draw [arrow] (total.south) -- (experts);
        \draw [arrow] (friends.south) -- (price);
        \draw [arrow] (friends.south) -- (fuel);
        \draw [arrow] (friends.south) -- (safety);
        \draw [arrow] (experts.south) -- (price);
        \draw [arrow] (experts.south) -- (fuel);
        \draw [arrow] (experts.south) -- (safety);
        \draw [arrow] (price.south) -- (honda);
        \draw [arrow] (price.south) -- (bmw);
        \draw [arrow] (fuel.south) -- (honda);
        \draw [arrow] (fuel.south) -- (bmw);
        \draw [arrow] (safety.south) -- (honda);
        \draw [arrow] (safety.south) -- (bmw);
        \end{tikzpicture}
    \caption{AHP model for the car selection problem}
    \label{AHP_Car_Selection}
\end{figure}
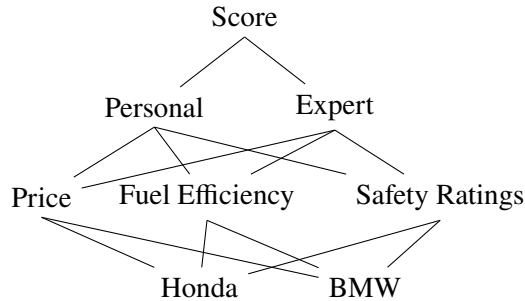

\paragraph{Step 1. Decomposition} The attributes identified for a problem are organized in a hierarchy. For example, Figure \ref{AHP_Car_Selection} shows the decomposition step applied to the car selection problem, which takes into account the personal preferences as well as expert evaluations. Each quantify their preference for a car by specifying its various attributes which are price, fuel efficiency, and safety ratings. (To make the example more manageable, we ignore the comfort attribute in the following and consider only the Honda and BMW alternatives.)

The bottom nodes represent the two alternatives from which one is to be selected. Each node on the level above (such as Price) is an attribute for each of the alternatives, and its value is given in a corresponding matrix that captures the relationship between the two levels of the hierarchies.
The AHP model is essentially a tree, represented as a DAG that shares common children.

\newcommand{\adv}{\ensuremath{A}}
\newcommand{\featadv}{\ensuremath{F}}
\newcommand{\carfeat}{\ensuremath{C}}

\begin{figure}[t]
\begin{small}
    \begingroup
    \setlength{\tabcolsep}{5pt} 
    \renewcommand{\arraystretch}{1.2} 
\begin{center}
{\small
    \begin{tabular}[t]{l|c}
    \multicolumn{1}{c|}{$B_1=\adv$}   
    & \catfmt{Car} \\
    \hline  
    \friends  & .6 \\ 
    \experts  & .4
    \end{tabular}
\ \qquad
    \begin{tabular}[t]{l|cc}
    \multicolumn{1}{c|}{$B_2=\featadv$}   
    & \friends & \experts   \\
    \hline  
    \price  & .5 & .2   \\ 
    \fuel   & .3 & .4  \\ 
    \safety  & .2 & .4  
    \end{tabular}
\ \qquad
    \begin{tabular}[t]{l|ccc}
    \multicolumn{1}{c|}{$B_3=\carfeat$}    
    & \price & \fuel & \safety  \\
    \hline  
    \honda  & .6 & .6 & .3  \\ 
    \bmw    & .4 & .4 & .7 
    \end{tabular}
}
\end{center}
\endgroup
\end{small}
\caption{Car selection decision matrices. \adv: weight of advice, \featadv: feature advice, \carfeat: car features.}
\label{fig:AHPExampleMatrices}
\end{figure}

\paragraph{Step 2. Comparative Judgment} The second step generates a matrix of pair-wise comparisons of all attributes in a level with respect to each related attribute in the level immediately above it. 
For a DAG of $n$ levels, $B_i$ for $(1\leq i < n)$ is the comparison matrix for the elements at level $i$ and $i+1$. 
The comparison matrices for our example $B_1=\adv$, $B_2=\featadv$, and $B_3=\carfeat$ are shown in Figure \ref{fig:AHPExampleMatrices}. 
(Note that we have normalized the numbers to account for the AHP constraint that the numbers in each column must sum up to 1. Moreover, we changed the numbers slightly to make the example a bit more interesting; we also rounded them to make the subsequent arithmetic easier to follow.)

Consider the matrix $B_3$. Since the level 4 of the DAG consists of 2 nodes, and level 3 consists of 3 nodes, the dimension of matrix $B_3$ is $2\times 3$. This matrix relates each feature with each car.
Specifically, the entry .6 for \fuel\ and \honda\ in $C$ says that a Honda's fuel efficiency is considered to be 50\% better than a BMW's.

\paragraph{Step 3. Synthesis of priorities} This step generates the global (or composite) priorities of the elements at the lowest level of the hierarchy. 
Given the priority matrices $B_1,\ldots,B_n$, the priority vector $W$ corresponding to the alternatives at the leaf nodes of the DAG is given by the matrix product
$W = B_n  B_{n-1} \ldots B_2 B_1$.
In our example, synthesis of priorities produces the following result.
%
%
\[
W = B_3 B_2 B_1 = \carfeat\featadv\adv =
\left[
\begin{array}{l} 0.516 \\ 0.484 \end{array}
\right]
\begin{tabular}{l} 
\small{\text{Honda}} \\
\small{\text{BMW}} 
\end{tabular}
\]
The synthesis concludes that Honda is the best car to buy, since it has a slight edge over BMW. 

This is all well and good, but since the scores for the two alternatives are quite close, the decision maker might be interested to know \emph{why} Honda is better and which assumptions lead to this conclusion.
An answer to this question is not obvious due to the complicated hierarchical relationships between the various nodes in the corresponding AHP diagram.
In the following, we demonstrate how to generate explanations for the linear and hierarchical MCDM techniques.

\section{Explaining Decisions of Linear MCDM With Value Decompositions}
\label{sec:theory}
\label{sec:vd}

%
Consider again the car selection example from Figure \ref{fig:ScoreCar}. We abbreviate the alternatives Toyota, Honda, and BMW by $T$, $H$, and $B$, respectively. The normalized contributions for the categories price (\price), fuel efficiency (\fuel), safety rating (\safety), and comfort (\comfort) are combined into a total score for every car alternative.
This view can be formalized using the concepts of \emph{value decomposition} and \emph{valuation}.
Given a set of categories \categories, a mapping $v:\categories\to\mathbb{R}$ is called a \emph{value decomposition} (with respect to \categories). The (total) \emph{value} of a value decomposition is defined as the sum of its components, that is, $\total{v}=\sum_{(c,x)\in v}x$.
A \emph{valuation} for a set $S$ (with respect to \categories) is a function \valSym\ that maps each element of $S$ to a corresponding value decomposition, that is, $\valSym: S\to (\categories\to\mathbb{R})$ (or, $\valSym: S\to \mathbb{R}^\categories$). We write $\tval{A}$ to denote the total value of $A$'s value decomposition.

The value decompositions for the alternatives in our example can be derived from Figure \ref{fig:ScoreCar} by multiplying the scores in each column by the weighting factor for that column, see Figure \ref{fig:DecompCar}.

\begin{figure}[t]
    \centering
    \begin{small}
    \begingroup
    \setlength{\tabcolsep}{5pt} 
    \begin{tabular}{r|cccc|c}
     & \price & \fuel & \safety  &  \comfort &  $\tau$ \\[2pt]
    \hline  
    \smallbf{Toyota}  &  0.10 &	0.34 & 0.27 & 0.14 & 0.85 \\[2pt] 
    \smallbf{Honda}   &  0.09 &	0.40 & 0.23 & 0.17 & 0.89 \\[2pt]
    \smallbf{BMW}     &  0.08 & 0.37 & 0.30 & 0.20 & 0.95
    \end{tabular}
    \endgroup
    \end{small}
    \caption{Value decompositions for the car selection example}
    \label{fig:DecompCar}
\end{figure}

To explain why BMW is the best choice, we have to specifically explain why it was chosen over Honda, the second-best alternative. We can also explain why BWM was chosen over Toyota, but that decision is not as close and therefore not as much in need of an explanation. We therefore focus on comparing BMW with Honda.

Focusing on $B$ and $H$ with their respective value decompositions
%
$v_B=\cardecomp{.08}{.37}{.30}{.20}$ and %
$v_H=\cardecomp{.09}{.40}{.23}{.17}$ leads to the valuation %
$\valSym = \{B\mapsto v_B, H\mapsto v_H\}$.
The elements of $S$ can be ordered based on the valuation totals in an obvious way: 
\[
\forall X,Y \in S.\ X>Y \Leftrightarrow \tval{X}>\tval{Y}
\]
When we ask why alternative $X$ was chosen over $Y$, the obvious explanation is to give the valuation totals, which provide the justification $\tval{X}>\tval{Y}$.
However, such an answer might not be useful, since it ignores the categories that link the raw numbers to the application domain and thus lacks a context to interpret the numbers.
In our example, BMW is chosen, since $\tval{B}=0.95 > \tval{H}=0.89$, which might be surprising because Honda is clearly cheaper as well as more fuel efficient.

If the value decomposition is maintained during the computation, we can generate a more detailed explanation.
First, we can rewrite $\tval{X}>\tval{Y}$ as $\tval{X}-\tval{Y}>0$, which suggests the definition of the \emph{valuation difference} between two elements $X$ and $Y$ as follows.
\[
\diff[\valSym]{X,Y} = \{(c,x-y)\ |\ (c,x)\in\val{X}\wedge(c,y)\in\val{Y}\}
\]
(In the following we will omit \valSym\ whenever it is clear from the context.)
The total of the valuation difference \tdiff{X,Y} is given by the sum of all components, just like the total of a value decomposition.
In our example we have 
$\diff{B,H} = \cardecomp{-.01}{-.03}{.07}{.03}$.
It is clear that the valuation difference generally contains positive and negative entries and that for $\diff{X,Y}>0$ to hold, the sum of the positive entries must exceed the absolute value of the sum of the negative entries.
We call the negative components of a valuation difference its \emph{barrier}. It is defined as follows.
\[
\barr[\valSym]{X,Y}=\{(c,x)\ |\ (c,x)\in \diff[\valSym]{X,Y}\wedge x<0\}
\]
The total value \tbarr{X,Y} is again the sum of all the components.
In our example we have $\barr{B,H} = \{\price\mapsto-0.01,\fuel\mapsto-.03\}$ and $\tbarr{B,H} = -.04$.
The decision to select $X$ over $Y$ needs as support some, but not necessarily all, of the positive components of \diff{X,Y}, which are called the \emph{dominator candidates} and which are defined as follows.
%
\[
\resol[\valSym]{X,Y}=\{(c,x)\ |\ (c,x)\in \diff[\valSym]{X,Y}\wedge x>0\}
\]
Any subset of \resol{X,Y} whose total is larger than $|\tbarr{X,Y}|$ will suffice as an explanation. We call such a subset a \emph{dominator}. The set of all dominators is defined as follows.
%
\[
\domi[\valSym]{X,Y}=\{D\ |\ D\subseteq\resol[\valSym]{X,Y}\wedge \total{D}>|\tbarr{X,Y}|\}
\]
In our example we have two dominators, that is,
$\domi{B,H} = \{\{\safety\mapsto .07\}, \{\safety\mapsto .07,\comfort \mapsto .03\}\}$.
The smaller a dominator, the better it is suited as an explanation, since it requires fewer details to explain how the barrier is overcome. We therefore define a \emph{minimal dominating set} (MDS) as any dominator with the fewest possible number of dominator candidates.
\[
\mdomi[\valSym]{X,Y}=\{D\ |\ D\subseteq\domi[\valSym]{X,Y}\wedge D'\subset D\Rightarrow D'\notin \domi{X,Y}\}
\]
Note that \mdomiSym\ may contain multiple elements, which means that minimal dominators are in general not unique. In other words, a decision may have different minimally sized explanations. 
In our example, the only MDS is $\mdomi{B,H}=\{\safety\mapsto .07\}$; it captures the explanation that BMW is to be preferred over Honda due to the significant difference in the safety ratings of the two cars alone; we don't have to mention comfort at all to explain the decision. 

We can apply the described technique to WPM examples by simply using multiplication for aggregation (using the multiplicative identity 1) and division for computing valuation differences. The barrier set then consists of all the components with values less than 1, and the dominator set consists of components with values greater than 1. An MDS explanation in this case is the smallest subset of dominator components whose product of component values will exceed that of the inverse of product of all the component values in the barrier set.
Alternatively, we could apply the log transform to the individual component values and then use the additive version of MDS.

\section{Hierarchical Value Decomposition} 
\label{sec:theoryAHP}

The idea of value decomposition relies on the fact that each alternative consists of a flat list of attributes. 
In contrast, attributes in the AHP setting are recursively decomposed into sub-attributes forming a hierarchical structure, which raises the question of whether the idea of MDS-based explanations can also work for hierarchical decision-making methods. This would require extending the concepts of value decomposition and dominators meaningfully to the hierarchical case.
%

To this end, we define the concept of a \emph{hierarchical value decomposition}, which records the individual contributions of the attributes at the various levels towards the overall priority of an alternative. The hierarchical value decomposition for a priority value of an alternative in AHP results in a tree that maps attributes to values. Since the synthesis of priorities in AHP consists of multiplication of decision matrices, we need a way to trace this matrix multiplication to come up with the tree structure, which we describe in this section. This tree forms the basis of explanations in the hierarchical case, which we describe in Section \ref{sec:AHP_Explain}.

\newcommand{\msymb}{\ensuremath{\cdot}}

\subsection{Tracing Matrix Multiplication with Value Decomposition Trees}\label{sec:vdTree}

Consider the decision matrix $B_3=\carfeat$ from our example. Multiplying \carfeat\ with $B_2=\featadv$ yields a $2\times 2$ matrix, where each element is the sum of 3 products.
\[
  \begin{bmatrix}
    .6 ~&~ .6 ~&~  .3\\
    .4 ~&~ .4 ~&~  .7\\
  \end{bmatrix}
  \begin{bmatrix}
    .5 ~&~ .2\\
    .3 ~&~ .4 \\
    .2 ~&~ .4 
  \end{bmatrix} = 
\begin{bmatrix}
    .6\msymb .5 + .6\msymb .3 + .3\msymb .2 ~&~ .6\msymb .2 + .6\msymb .4 + .3\msymb .4 \\
     \ldots ~&~ \ldots
  \end{bmatrix}
\]
Any such sum of products can be visually represented as a tree whose leaf nodes contain the products and whose internal nodes contain the sum of the values of its children. 
Here are the trees for the elements of the first row of the resulting matrix.

\bigskip
\begin{center}
\begin{tikzpicture}[scale=1.00]
    \tikzstyle{level 1}=[sibling distance=15mm]
\node {$.6\msymb .5 + .6\msymb .3 + .3\msymb .2$} [->]
      child {node {$.6\msymb .5$}}
      child {node {$.6\msymb .3$}}
      child {node {$.3\msymb .2$}};
\end{tikzpicture}
\qquad
\begin{tikzpicture}[scale=1.00]
    \tikzstyle{level 1}=[sibling distance=15mm]
\node {$.6\msymb .2 + .6\msymb .4 + .3\msymb .4$} [->]
      child {node {$.6\msymb .2$}}
      child {node {$.6\msymb .4$}}
      child {node {$.3\msymb .4$}};
\end{tikzpicture}
\end{center}
In the next step we multiply the result of $B_3 B_2=\carfeat\featadv$ with the first decision matrix $B_1=\adv$, which means to multiply each summand of each matrix element with another factor and creating new sums of the results.
\[
  B_3 B_2 B_1 = \carfeat\featadv\adv = 
  \begin{bmatrix}
    .6\msymb .5 + .6\msymb .3 + .3\msymb .2 ~&~ .6\msymb .2 + .6\msymb .4 + .3\msymb .4 \\
   \ldots ~&~ \ldots
  \end{bmatrix}
  \begin{bmatrix}
    .6 \\
    .4 \\
  \end{bmatrix}
\]
The first entry of the resulting vector is given by the following value.
\[
\begin{array}{@{}r@{\ }l@{\ +\ }l}
&
(.6\msymb .5 + .6\msymb .3 + .3\msymb .2)\msymb .6 &
(.6\msymb .2 + .6\msymb .4 + .3\msymb .4)\msymb .4
\\
=&
(.6\msymb .5\msymb .6 + .6\msymb .3\msymb .6 + .3\msymb .2\msymb .6) &
(.6\msymb .2\msymb .4 + .6\msymb .4\msymb .4 + .3\msymb .4\msymb .4)
\end{array}
\]
Again, this sum of products can be represented by a tree, now with three levels.

\bigskip
\begin{center}
\begin{tikzpicture}[scale=1.00]
    \tikzstyle{level 1}=[sibling distance=65mm]
    \tikzstyle{level 2}=[sibling distance=21mm]
\node {$.6\msymb .5\msymb .6 + .6\msymb .3\msymb .6 + .3\msymb .2\msymb .6+
        .6\msymb .2\msymb .4 + .6\msymb .4\msymb .4 + .3\msymb .4\msymb .4$} [->]
child{node {$.6\msymb .5\msymb .6 + .6\msymb .3\msymb .6 + .3\msymb .2\msymb .6$}
      child {node {$.6\msymb .5\msymb .6$}}
      child {node {$.6\msymb .3\msymb .6$}}
      child {node {$.3\msymb .2\msymb .6$}}
}
child{node {$.6\msymb .2\msymb .4 + .6\msymb .4\msymb .4 + .3\msymb .4\msymb .4$}
      child {node {$.6\msymb .2\msymb .4$}}
      child {node {$.6\msymb .3\msymb .4$}}
      child {node {$.3\msymb .3\msymb .4$}}
}
;
\end{tikzpicture}
\end{center}
This tree represents the contribution of the various attributes toward the overall priority value of Honda (.516). For example, the left subtree of the root represents the contribution of the personal opinion. Specifically, the root of the left subtree contains the total of the personal  opinion (.324), whereas the children contain the decomposition of that value into the individual values for price (.180), fuel efficiency (.108), and safety ratings (.036) of the personal opinion.

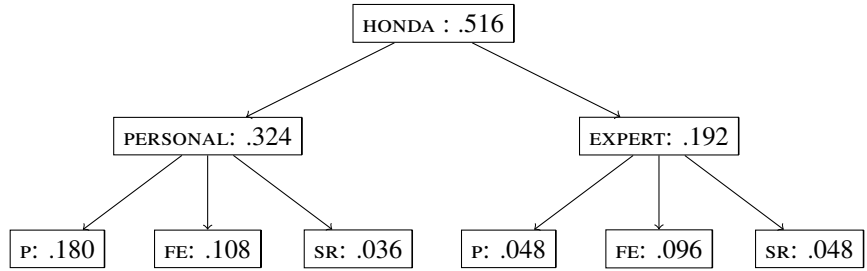
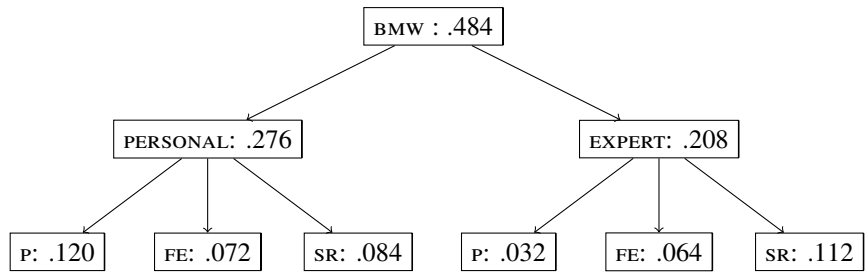
\begin{figure}
%
\begin{subfigure}[b]{\textwidth}
\begin{center}
\vspace{1em}
\begin{tikzpicture}[scale=1.00]
    \tikzstyle{bplus}=[rectangle split, rectangle split horizontal, rectangle split parts=6,
                      rectangle split ignore empty parts,draw, fill=white, font =\small]
    \tikzstyle{every node}=[bplus]
    \tikzstyle{level 1}=[sibling distance=60mm]
    \tikzstyle{level 2}=[sibling distance=20mm]
    \tikzstyle{arrow}=[thick]
\node {$\honda: .516$} [->]
  child {node{\friends: $.324$}
                  child {node {\price: $.180$ }}
                  child {node {\fuel: $.108$}}
                  child {node {\safety: $.036$}}
        }
  child {node{\experts: $.192$}
                  child {node {\price: $.048$ }}
                  child {node {\fuel: $.096$}}
                  child {node {\safety: $.048$}}
        }
;
\end{tikzpicture}
\end{center}
    \caption{VD tree $\valSym_\honda$ showing the attribute contributions toward the overall priority of Honda}
    \label{fig:vdtreehonda}
\end{subfigure}

\begin{subfigure}[b]{\textwidth}
\begin{center}
\vspace{1em}
\begin{tikzpicture}[scale=1.00]
    \tikzstyle{bplus}=[rectangle split, rectangle split horizontal, rectangle split parts=6,
                      rectangle split ignore empty parts,draw, fill=white, font =\small]
    \tikzstyle{every node}=[bplus]
    \tikzstyle{level 1}=[sibling distance=60mm]
    \tikzstyle{level 2}=[sibling distance=20mm]
    \tikzstyle{arrow}=[thick]
\node {$\bmw: .484$} [->]
  child {node{\friends: $.276$}
                  child {node {\price: $.120$ }}
                  child {node {\fuel: $.072$}}
                  child {node {\safety: $.084$}}
        }
  child {node{\experts: $.208$}
                  child {node {\price: $.032$ }}
                  child {node {\fuel: $.064$}}
                  child {node {\safety: $.112$}}
        }
;
\end{tikzpicture}
\end{center}
    \caption{VD tree $\valSym_\bmw$ showing the attribute contributions toward the overall priority of BMW}
    \label{fig:vdtreebmw}
\end{subfigure}

\caption{Value decomposition trees for the hierarchical car example}
    \label{fig:vdtree}
\end{figure}

\newcommand{\ri}[1]{\ensuremath{\omega_{#1}}}

To assign meaning to the tree components, we can label them with the attribute names, which are already used as row and column labels, linking the individual and aggregated value contributions to the attributes of the decision problem.
We can observe the following. 

\begin{enumerate}[(A)]
\item The row labels of $B_n$ should label the roots of the trees for the resulting priority vector.
\item The column labels of matrix $B_\ell$ (for $1<\ell\leq n$), which are equal to the row labels of matrix $B_{\ell-1}$, should label the nodes on level $\ell$.
\end{enumerate}
We call each such labeled tree a \emph{value decomposition tree}, or \emph{VD tree} for short. An example is shown in Figure \ref{fig:vdtreehonda}.

Now we describe a simple method to create VD trees from a sequence of matrices $B_1,\ldots,B_n$.
%
Observations (A) and (B) tell us that every path from root to a leaf in a VD tree is labeled by row labels taken from matrices in the order $B_n,B_1,\ldots,B_{n-1}$. Let's write $[\ri{n}, \ri{1}, \ldots, \ri{n-1}]$ for the row indices corresponding to those labels.
For example, in Figure \ref{fig:vdtreehonda} the leftmost leaf is identified by the path of row labels $(\honda,\friends,\price)$, which corresponds to the row indices $(1,1,1)$, and the rightmost leaf is identified by the path $(\honda,\experts,\safety)$, which corresponds to the row indices $(1,2,3)$.
The values in a VD tree are determined as follows.

\begin{itemize}
\item[\textbf{1.}]
Assign each leaf connected to the root a path of row labels with indices $[\ri{n}, \ri{1}, \ldots, \ri{n-1}]$ the value
$B_n[\ri{n},\ri{n-1}] \times
B_{n-1}[\ri{n-1},\ri{n-2}] \times
\ldots \times
B_1[\ri{1}]
$.
\item[\textbf{2.}]
Assign each internal node the sum of the values of its children.
\end{itemize}
%
%
%
%
In the example from Figure \ref{fig:vdtreehonda}, the value of the leftmost leaf is therefore computed as $B_3[1,1] \times B_2[1,1] \times B_1[1] = .6 \times .5 \times .6 = .180$.
Similarly, the value of the rightmost leaf is computed as
$B_3[1,3] \times B_2[3,2] \times B_1[2] = .3 \times .4 \times .4 = .048$.
%
The sums of the internal nodes and the root are computed in the obvious way.

\newcommand{\vdtreeSym}{\textit{V}}
\newcommand{\vdnode}[2]{\ensuremath{#1\mathord:\,#2}}
\newcommand{\vdtree}[3]{\ensuremath{\vdnode{#1}{#2}[#3]}}
\newcommand{\rootx}[1]{\ensuremath{\rho(#1)}}
\newcommand{\lkup}[2][\vdtreeSym]{\ensuremath{#1(#2)}}
\newcommand{\rootval}[1]{\ensuremath{|#1|}}
\newcommand{\chld}[1]{\ensuremath{\gamma(#1)}}

\subsection{Hierarchical Valuation Differences}
\label{sec:AHP_Explain}

To generate explanations from VD trees we have to generalize the concepts of valuation difference to the hierarchical case. 
%
%
%

%
%

The definition of valuation changes only slightly insofar as elements of the set $S$ (which are identical to the row labels of $B_n$) are mapped to VD trees instead of plain value decompositions. For our car example the valuation is $\valSym = \{\bmw\mapsto \vdtreeSym_\bmw, \honda\mapsto \vdtreeSym_\honda\}$ (cf.\ Figure \ref{fig:vdtree}). 

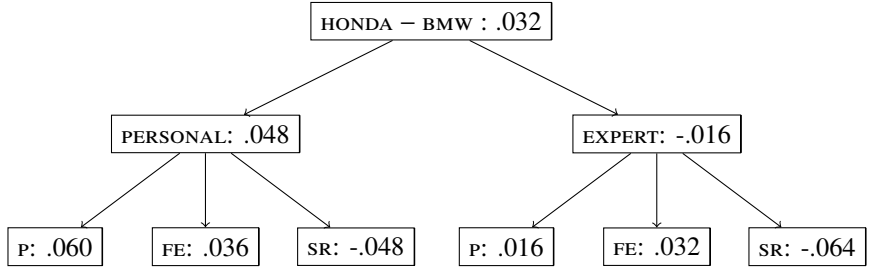
\begin{figure}
\begin{center}
\vspace{1em}
\begin{tikzpicture}[scale=1.00]
    \tikzstyle{bplus}=[rectangle split, rectangle split horizontal, rectangle split parts=6,
                      rectangle split ignore empty parts,draw, fill=white, font =\small]
    \tikzstyle{every node}=[bplus]
    \tikzstyle{level 1}=[sibling distance=60mm]
    \tikzstyle{level 2}=[sibling distance=20mm]
    \tikzstyle{arrow}=[thick]
\node {$\honda - \bmw: .032$} [->]
  child {node{\friends: $.048$}
                  child {node {\price: $.060$ }}
                  child {node {\fuel: $.036$}}
                  child {node {\safety: -$.048$}}
        }
  child {node{\experts: -$.016$}
                  child {node {\price: $.016$ }}
                  child {node {\fuel: $.032$}}
                  child {node {\safety: -$.064$}}
        }
;
\end{tikzpicture}
\end{center}
    \caption{Hierarchical valuation difference between Honda and BMW, represented as a VD tree $\vdtreeSym_{\honda-\bmw}$.}
    \label{fig:vdtreehondabmw}
\end{figure}

The concept of valuation difference then extends in a natural way to the hierarchical case.
First, we write $V(\ell)$ for the value in the node that is identified by the path of labels $\ell$ from the root.
Then the valuation difference \diff[\valSym]{A,B} between two VD trees $V_A$ and $V_B$ is defined as the VD tree $\vdtreeSym_{A-B}$ which has the same structure and labels as $\vdtreeSym_A$ and $\vdtreeSym_B$ (except for the root label) such that for all root-path labels $\ell$ in $\vdtreeSym_A$ except the root:
$\lkup[\vdtreeSym_{A-B}]{\ell}=\lkup[\vdtreeSym_A]{\ell}-\lkup[\vdtreeSym_B]{\ell}$.
%
The label of the root of $\vdtreeSym_{A-B}$ is $A-B$, and the value of the root is  $\vdtreeSym(A)-\vdtreeSym(B)$. As an example, the VD tree $\vdtreeSym_{\honda-\bmw}$ is shown in Figure \ref{fig:vdtreehondabmw}. 

\section{Hierarchical Explanations With Value Decomposition Trees}\label{sec:Hierarchical_explainataion}

We saw in Section \ref{sec:theoryAHP} that the children of any node in a VD tree represent a value decomposition for that node, which means that the leaves of a VD tree represent the most granular value decomposition for the difference between scores of the alternatives. The component value for such a value decomposition is the value at a given leaf node, and the component label is the list of node labels on the path from the root to the leaf node. Since the root node label is shared across all decompositions, we can usually remove it without losing any relevant information. 
Moreover, by construction all non-leaf nodes in a VD trees represent redundant information, that is, the aggregation of the values of their children. Therefore, each VD tree can be succinctly represented by its root-path-labeled leaves.
For our example, we have for $\vdtreeSym_{\honda-\bmw} = \diff{\honda,\bmw}$:
\begin{align*}
\vdtreeSym_{\honda-\bmw} \cong \{ 
 &(\personal,\price)\mapsto .060,(\personal,\fuel)\mapsto .036,(\personal,\safety)\mapsto-.048,\\  
 & (\experts,\price)\mapsto.016,(\experts,\fuel)\mapsto .032,(\experts,\safety)\mapsto -.064\}\\
\end{align*}
To this flat mapping representation of hierarchical value decompositions we can now apply the concepts of dominators and MDS as defined in Section \ref{sec:vd}. %
For example, the barrier is given by the negative components $(\personal,\fuel)$ and $(\experts,\fuel)$.
\begin{align*}
    & \barr{\honda,\bmw} = \{(\personal,\safety) \mapsto -.048, (\experts,\safety)\mapsto -.064\} \text{ and } \\
    & \tbarr{\honda,\bmw} = -1.12. 
\end{align*}
That is, BMW has an advantage over Honda in the personal and expert opinion about safety ratings. We can justify Honda as the preferred car with any dominating set, that is, any set of components whose sum exceeds the absolute value of the barrier. Here we have two dominators.
\[
\begin{array}{l}
\domi{\honda,\bmw} = \\
\{ \{(\personal,\price) \mapsto .060,(\personal,\fuel) \mapsto .036, (\experts,\fuel) \mapsto .032\}, 
\\ 
\ \{(\personal,\price) \mapsto .060,(\personal,\fuel) \mapsto .036, (\experts,\price) \mapsto .016, (\experts,\fuel) \mapsto .032\}\} \\ 
\mdomi{\honda,\bmw} =  \{(\personal,\price)\mapsto .060,(\personal,\fuel)\mapsto .036,(\experts,\fuel) \mapsto .032 \}
\end{array}
\]

\par\medskip\noindent
It is obvious that the first dominator is the MDS in this case, since it is a proper subset of the second one. 
Interpreted as an explanation, the MDS says that personal preference for Honda's cheaper price as well as personal and expert favorable opinion for its fuel efficiency more than compensates for BMW's advantage in safety rating, making Honda the preferred car overall.

\section{Explanation Simplification}
\label{sec:coarsegrained}

Consider the MDS component $(\personal,\price)\mapsto .060$. Its attribute labels \personal\ and \price\ come from different levels of a VD tree. Comprehending such an explanation can be challenging, especially as the number of levels increases. A simplified explanation that employs labels from just one level might be easier to understand and thus may have more explanatory value, even if it is less specific.

The hierarchical decomposition of valuation differences makes it possible to provide explanations on different levels of the VD tree. In particular, the tree structure enables decision makers to inquire specific details about the reasons for a decision at the various levels of the VD tree.
%

For example in the car selection decision an answer to the question ``Why is Honda the preferred option with respect to the decision makers?'' is given by the children of $\vdtreeSym_{\honda-\bmw}$ in Figure \ref{fig:vdtreehondabmw}: The value -.016 for \experts\ represents a barrier, and the value .048 for \friends\ is the MDS, which corresponds to the explanation that the positive personal opinion of Honda outweighs the negative opinion of the experts. This explanation mentions only two values and is simpler, albeit less specific, than the explanation given in the previous section.

Similarly, we could ask ``Why is Honda the preferred option with respect to the features?''. The answer is given by the children of $\vdtreeSym_{\friends}$ and $\vdtreeSym_{\experts}$ in $\vdtreeSym_{\honda-\bmw}$: The overall value for \safety\ (-0.064 + -0.048 = -0.112), obtained by summing the \safety\ values of $\vdtreeSym_{\friends}$ and $\vdtreeSym_{\experts}$, represents a barrier and the total values for \price\ (0.060 + 0.016 = 0.076) and \fuel\ (0.036 + 0.032 = 0.068) obtained by summing the corresponding \price\ and \fuel\ values of $\vdtreeSym_{\friends}$ and  $\vdtreeSym_{\experts}$ is the MDS. This corresponds to the explanation that although the safety ratings are against Honda, those are more than compensated by a better price and fuel efficiency. Note that an MDS for level $\ell+1$ is \emph{not} a refinement of the MDS for level $\ell$; rather they are independent explanations for the same outcome. 

\newcommand{\level}[2][\ell]{\ensuremath{\underline{#2}_{#1}}}
\renewcommand{\leveled}[2][\ell]{\ensuremath{#2_{#1}}}
\newcommand{\agg}[1]{\ensuremath{#1^\cup}}
\newcommand{\aggchld}[1]{\ensuremath{\agg{\gamma}(#1)}}
\newcommand{\negn}{\textrm{-}}

\newcommand{\fagg}[2]{\ensuremath{#1/#2}}
\newcommand{\proj}[1]{\ensuremath{\pi_{#1}}}
\newcommand{\pagg}[2]{\ensuremath{#1/\proj{#2}}}

To formalize the focusing on different levels in a VD tree, we need an operation for aggregating
functions over multiple domain values.
Specifically, given $f:A\to \mathbb{R}$ and $g: A\to B$, the \emph{aggregation} of $f$ with respect to $g$ is the function $\fagg{f}{g} : B\to \mathbb{R}$, defined as follows.
\[
\fagg{f}{g} = \{(x',\sum \{y\ |\ (x,y)\in f, g(x)=x'\})\ |\ x'\in g(A) \}
\]
We can use this aggregation to create mappings that summarize the values of a VD tree on different levels. Let \proj{n} be the function that selects (or projects onto) the $n$th element of a list or tuple. Then \fagg{V}{\proj{n}} creates an aggregation of the (root-path-labeled representation of the) VD tree that maps the labels on the $n$th level to their aggregated values.


%
For example, the levels 2 and 3 of the VD tree $\vdtreeSym_\honda$ can be obtained as follows.
\begin{align*}
\pagg{\vdtreeSym_{\honda}}{2} &= \cardecompAHPL{.324}{.192} \\ 
\pagg{\vdtreeSym_{\honda}}{3} &= \cardecompAHP{.180 + .048}{.108+.096}{.036+.048} \\ 
    & = \cardecompAHP{.228}{.204}{.084}
\end{align*}
Similarly, we can focus on different levels of a VD tree that stores valuation differences, and we can also focus the definitions of barrier, MDS, etc.\ by applying the corresponding function to the focused valuation difference.
\begin{align*}
\lbarr[\valSym]{A,B}  &= \barr[{\pagg{\diff[\valSym]{A,B}}{\ell}}]{A,B} &
\lresol[\valSym]{A,B} &= \resol[{\pagg{\diff[\valSym]{A,B}}{\ell}}]{A,B} \\
\ldomi[\valSym]{A,B}  &= \domi[{\pagg{\diff[\valSym]{A,B}}{\ell}}]{A,B} &
\lmdomi[\valSym]{A,B} &= \mdomi[{\pagg{\diff[\valSym]{A,B}}{\ell}}]{A,B}
\end{align*}
Applying these definitions to level 2, we get the following valuation difference, barrier and, MDS explanation, leading to the explanation we saw for level 2 at the beginning of this section.
\begin{align*}
    & \xdiff2{\honda,\bmw} = \{\personal \mapsto .048 , \experts \mapsto -.016 \} \\
    & \xbarr2{\honda,\bmw} = \{\personal \mapsto .048 \}\qquad 
    \xmdomi2{\honda,\bmw} = \{\experts \mapsto -.016 \}
\end{align*}
Similarly, we can compute these values for level 3 explanations. 
\begin{align*}
    & \xdiff3{\honda,\bmw} = \{\price \mapsto .076, \fuel \mapsto .068, \safety \mapsto -.112\} \\
    & \xbarr3{\honda,\bmw} = \{\price \mapsto .076, \fuel \mapsto .068 \}\qquad 
    \xmdomi3{\honda,\bmw} = \{\safety \mapsto -.112\}
\end{align*}

\section{Case Studies of Applications of MDS to AHP}
\label{sec:CaseStudy}

In this section we apply our explanation mechanism to two real-world AHP applications.

\subsection{Selecting Materials to Build Bridges in Rural Winsconsin Counties}

\newcommand{\reinforced}{\catfmt{R}}
\newcommand{\prestressed}{\catfmt{P}}
\newcommand{\steel}{\catfmt{S}}
\newcommand{\timber}{\catfmt{T}}
\newcommand{\statedot}{\catfmt{DOT}}
\newcommand{\consultants}{\catfmt{CONSULTANTS}}
\newcommand{\county}{\catfmt{OFFICIALS}}
\newcommand{\pastPerformance}{\catfmt{PP}}
\newcommand{\lifespan}{\catfmt{LS}}
\newcommand{\maintenance}{\catfmt{MN}}
\newcommand{\resistance}{\catfmt{RS}}
\newcommand{\initialcost}{\catfmt{IC}}
\newcommand{\lifecyclecost}{\catfmt{LC}}
\newcommand{\priorites}{\catfmt{IMPORTANCE}}

\begin{figure}[t]
    \centering
    \includegraphics[width=\columnwidth]{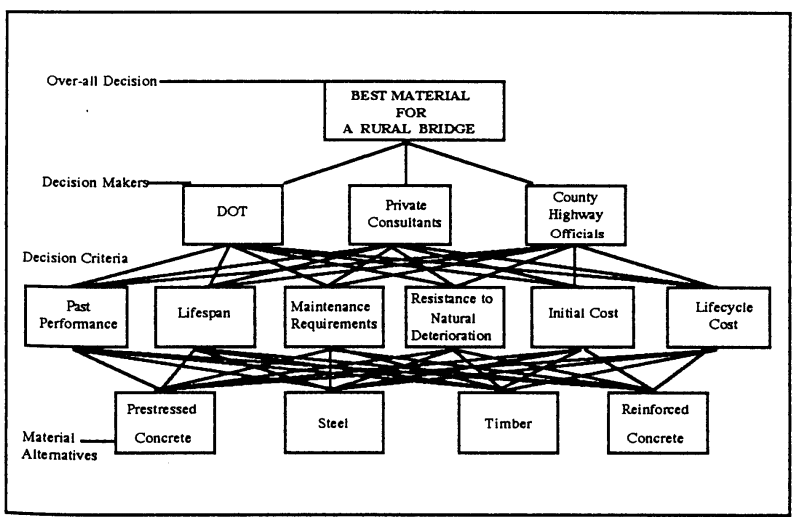}
    \caption{AHP model for selecting materials to build bridges. Figure taken from~\citet{articleBridgeAHP}.}
    \label{AHP_Bridge}
\end{figure}    

Figure \ref{AHP_Bridge} shows the AHP model for the problem of selecting the best material amongst prestressed concrete (\prestressed), steel (\steel), timber (\timber), and reinforced concrete (\reinforced) to build bridges in the rural counties of Wisconsin~\citep{articleBridgeAHP}.
The decision to select the best material takes into account the various stakeholders in the process (the state department of transport (\statedot), the private consultants (\consultants), and the county highway officials (\county)) who base their preferences of the materials on their characteristics such as past performance (\pastPerformance), lifespan (\lifespan), maintenance requirements (\maintenance), resistance to natural deterioration (\resistance), initial cost (\initialcost), and life cycle cost (\lifecyclecost). The decision matrices $B_1$, $B_2$, and $B_3$ for the problem are shown in Figure \ref{fig:AHPExampleMatricesBridge}. Note that the transposed matrix of $B_2$ is shown in the figure for easier presentation. 

\begin{figure}[t]
    \begingroup
    \setlength{\tabcolsep}{5pt} 
    \renewcommand{\arraystretch}{1.0} 
    \begin{tabular}{l|c}
    \multicolumn{1}{c|}{$B_1$} 
    & \priorites \\
    \hline  
    \statedot & .4 \\ 
    \consultants   & .2 \\ 
    \county  & .4
    \end{tabular}
\ \qquad
    \begin{tabular}{l|cccccc}
    \multicolumn{1}{c|}{$B^{\textrm{T}}_2$} 
    & \pastPerformance  & \lifespan & \maintenance & \resistance  &  \initialcost  & \lifecyclecost \\
    \hline  
    \statedot & .28 & .28 & .17 & .08 & .10 & .09 \\ 
    \consultants   & .08 & .08 & .35 & .08 & .32 & .09 \\ 
    \county  & .14 & .12 & .22 & .31 & .10 & .11\\
    \end{tabular}
    
\bigskip
    \begin{tabular}{l|cccccc}
    \multicolumn{1}{c|}{$B_3$} 
    & \pastPerformance & \lifespan & \maintenance  &  \resistance  & \initialcost & \lifecyclecost \\
    \hline  
    \prestressed & .33 & .27 & .42 & .32 & .23 & .28 \\ 
    \steel  & .09 & .16 & .08 & .09 & .15 & .09 \\ 
    \timber  & .20 & .23 & .23 & .26 & .32 & .31 \\
    \reinforced  & .38 & .34 & .27 & .33 & .29 & .31 \\
    \end{tabular}
    \endgroup
\caption{Decision matrices for the bridge material selection problem.}
\label{fig:AHPExampleMatricesBridge}
\end{figure}

The synthesis of priorities produces the following result for the various building materials and concludes that reinforced concrete is the best material to build the bridges, since it has a slightly higher priority value than the prestressed concrete.
\[
W = B_3 B_2 B_1 = 
\left[
\begin{array}{l} 0.319372 \\ 0.109007 \\ 0.251212 \\  0.320409 \end{array}
\right]
\begin{tabular}{l} 
\small{\text{Prestressed Concrete} (\prestressed)} \\
\small{\text{Steel} (\steel)} \\
\small{\text{Timber} (\timber)} \\
\small{\text{Reinforced Concrete} (\reinforced)}
\end{tabular}
\]
Since the \prestressed\  and \reinforced\ scores are very close, one might wonder why that is the case and which assumptions lead to this conclusion.

\begin{figure}[t]
\begin{center} 
\includegraphics[scale=1.00]{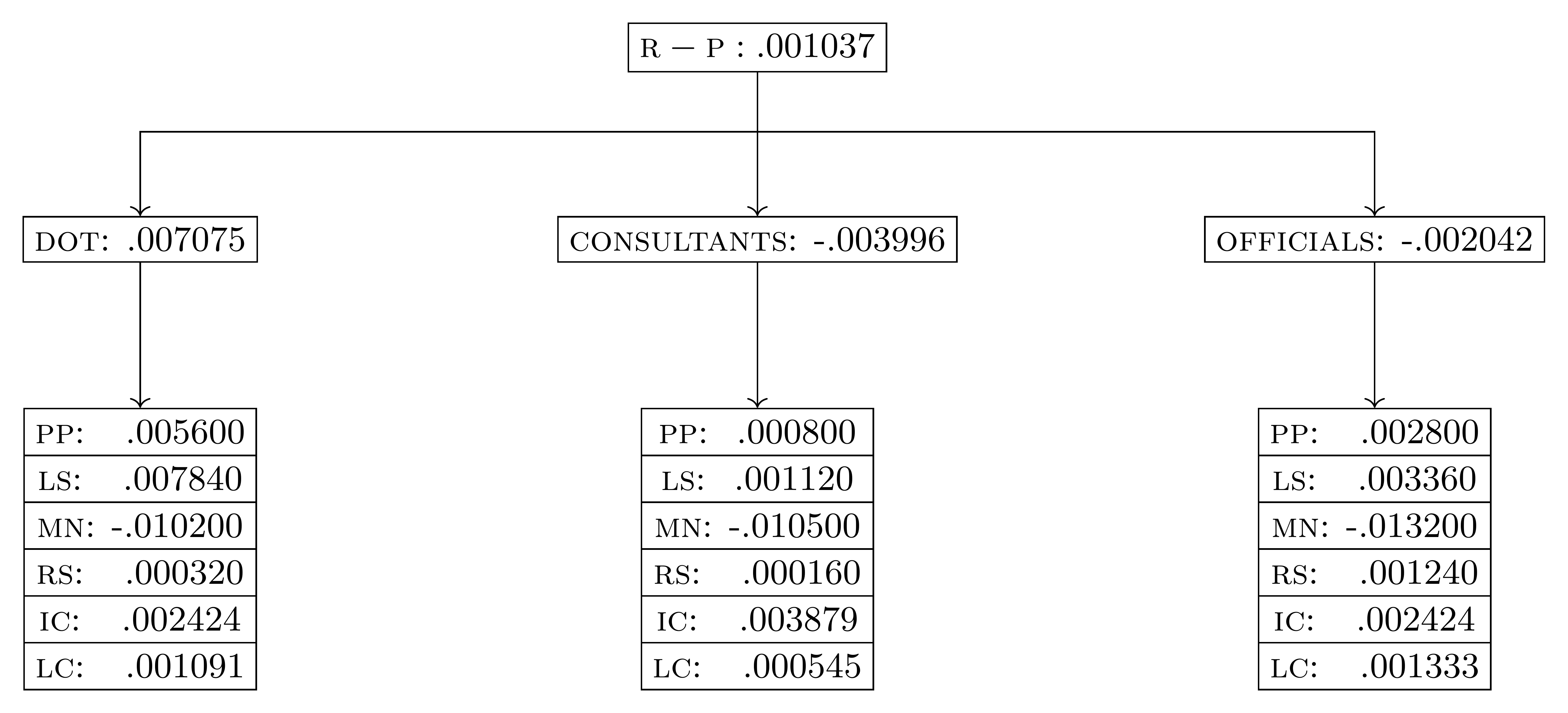}
\end{center}

\caption{Hierarchical valuation difference between \reinforced\ and \prestressed\ in the bridge selection example.}
    \label{fig:hvdtreeBridge}
\end{figure}

For lack of space, we show in Figure \ref{fig:hvdtreeBridge} only the hierarchical valuation difference (and not the individual VD trees) for the two best alternatives. From this we can generate an explanation for why reinforced concrete was preferred. The barrier comprises the opinions of the DOT, the consultants, and county officials with regard to the maintenance costs. 
\[
\barr{\reinforced,\prestressed}=
\{(\statedot,\maintenance) \mapsto -.0102, 
(\consultants,\maintenance) \mapsto -.0105, 
(\county,\maintenance) \mapsto -.0132\} 
\]
The barrier against the reinforced concrete is overcome with the following MDS explanation.
\begin{align*}
\mdomi{\reinforced,\prestressed} = 
 \{ &(\statedot,\pastPerformance) \mapsto .0056, (\statedot,\lifespan) \mapsto .00784, (\statedot,\initialcost) \mapsto  .002424, (\statedot, \lifecyclecost) \mapsto .001091, \\
&(\consultants,\pastPerformance) \mapsto .0008, (\consultants,\lifespan) \mapsto .00112, (\consultants,\initialcost)  \mapsto .003879, \\
& (\county,\pastPerformance) \mapsto .0028 ,   (\county,\lifespan) \mapsto .00336, (\county,\resistance) \mapsto .00124,\\ 
& (\county,\initialcost) \mapsto .002424, (\county, \lifecyclecost) \mapsto .001333
\}
\end{align*}
Since the priorities for reinforced and prestressed concrete are very close, the MDS contains a large number of components, which might be difficult to interpret.
We can help by generating single-level explanations, either in terms of the decision criteria or the decision makers. The valuation difference in terms of decision criteria is obtained by focusing on level 3. 
\[
\xdiff3{\reinforced,\prestressed} = \{\pastPerformance \mapsto .0092, \lifespan \mapsto .01232, \maintenance \mapsto -.0339, \resistance \mapsto .00172, \initialcost \mapsto .008727, \lifecyclecost \mapsto .00297\}
\]
In the valuation difference, maintenance is the only component acting as a barrier $\xbarr3{\reinforced,\prestressed} = \{\maintenance \mapsto -.0339\}$. The MDS explanation consists of the remaining components as shown below. 
\[
\xmdomi3{\reinforced,\prestressed} = 
 \{\pastPerformance \mapsto .0092, \lifespan \mapsto .01232, \resistance \mapsto .00172, \initialcost \mapsto .008727, \lifecyclecost \mapsto .00297\}
\]
The explanation at the level of decision criteria can be read like this: Although reinforced concrete has a disadvantage in terms of maintenance requirements, the cumulative advantage it has for the remaining criteria makes up for this disadvantage.

Similarly, we can get an explanation regarding decision makers by focusing on level 2. 
\[
\xdiff2{\reinforced,\prestressed} = \{\statedot \mapsto .007075, \consultants \mapsto -.003996, \county \mapsto -0.002042\}
\]
The consultants and county highway officials act as the barrier, but the MDS $\xdomi2{\reinforced,\prestressed} = \{\statedot \mapsto .007075\}$ tells us that the preference of the DOT compensates for this disadvantage.

\subsection{Supreme Court Rulings on Abortion}\label{sec:rovevwade}


\begin{figure}[t]
    \centering
    \includegraphics[scale=0.3]{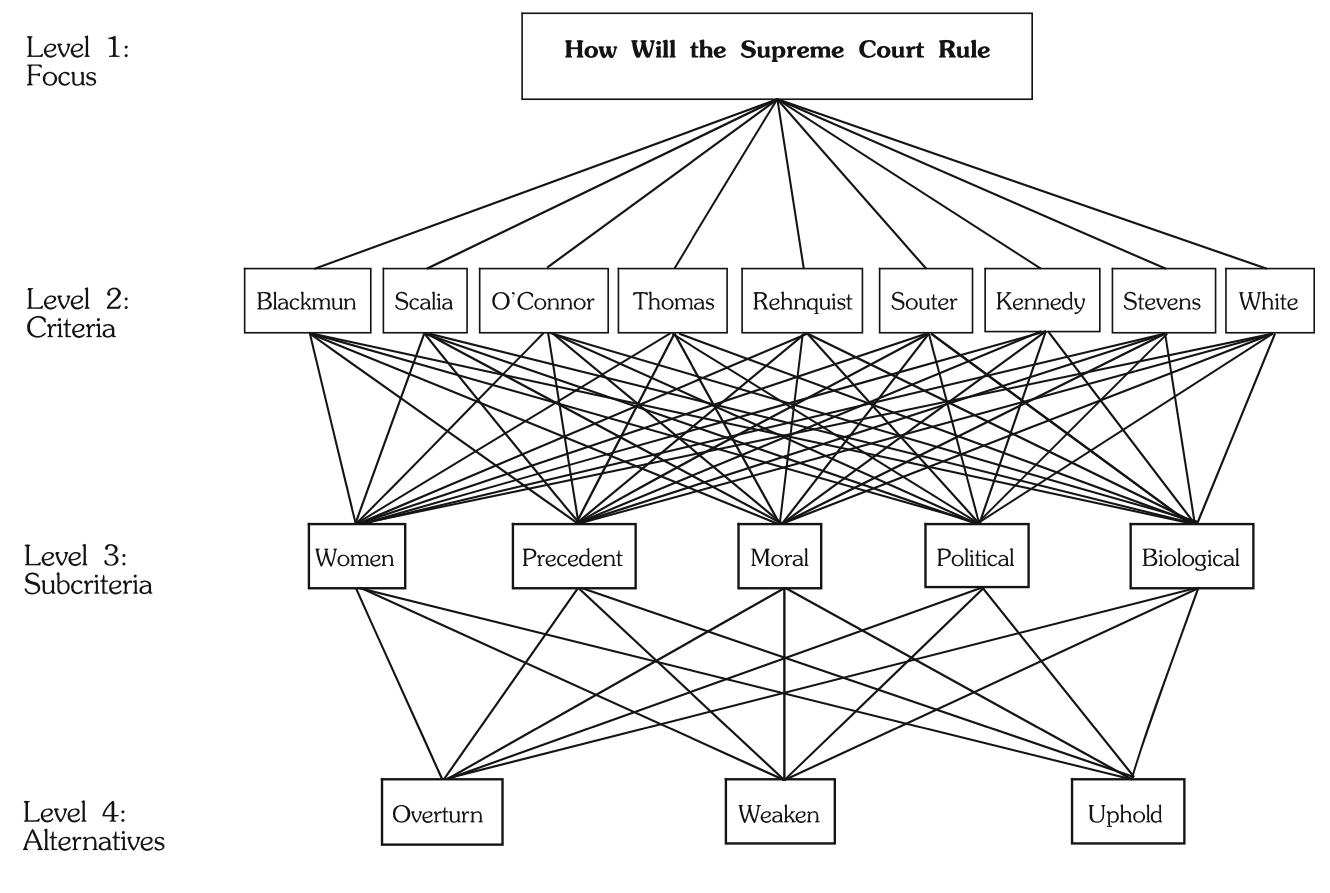}
    \caption{AHP model for predicting the Supreme Court decision on Pennsylvania abortion issue. Figure taken from~\cite{Saaty2012}.}
    \label{Court_Decision}
\end{figure}  

Roe v.\ Wade~\citep{roevswade} was a landmark decision of the U.S. Supreme Court in which the Court ruled that the Constitution of the United States generally protects a pregnant woman's liberty to choose to have an abortion. However, it was recently overturned by the Supreme Court, sparking off an intense public debate. It is in this context that the current AHP example~\citep{Saaty2012}, describing an AHP model to predict the Supreme Court ruling on a related issue of Roe v.\ Wade in 1992, and our techniques explaining why those outcomes were predicted gain special relevance. 
%

In the summer of 1992 the Supreme Court of the United States was supposed to rule on a controversial Pennsylvania statute restricting the rights of women in obtaining an abortion. Included in this statute were provisions requiring that doctors provide women with state-prescribed information about pregnancy and abortion, that the procedure be delayed 24 hours after the recitation, and that husbands be notified prior to the procedure. The lower court upheld the first two provisions, but declared unconstitutional the husband notification requirement. The AHP model for the example is shown in Figure \ref{Court_Decision}. It correctly predicted that the Supreme Court will uphold at least parts of the Pennsylvania statue and will, as a result, weaken the rights of women who choose to have an abortion in the state of Pennsylvania.

\newcommand{\women}{\catfmt{W}}
\newcommand{\precedent}{\catfmt{P}}
\newcommand{\moral}{\catfmt{M}}
\newcommand{\political}{\catfmt{O}}
\newcommand{\biological}{\catfmt{B}}
\newcommand{\overturn}{\catfmt{OVERTURN}}
\newcommand{\weaken}{\catfmt{WEAKEN}}

The model uses the nine Supreme Court justices as the criteria, giving each of them an equal weight. Beneath each justice there are five sub-criteria that were determined to be the most important for the judges to adjudicate on the matter. 

\begin{itemize}
    \item {\bf Women’s issues} (\women) These are issues deemed important by the pro-choice movement, such as the constitutional right of each woman to make her own decisions regarding her body. 
    \item {\bf Precedent} (\precedent) Cases that have gone before the Supreme Court since the early 1970s. 
    \item {\bf Moral issues} (\moral) Constitutional rights of the fetuses and the belief that abortion is murder. 
    \item {\bf Political issues} (\political) To make the decision-making process easier, the political issues are defined as conservatism. The original paper determined that conservatives are more pro-life than liberals. It also links Republicans with conservatism and Democrats with liberalism.
    \item {\bf Biological issues} (\biological) The medical concept of viability that specifies a certain time when the fetus is capable of independent survival outside the mother’s womb.
\end{itemize}

The AHP model envisaged three likely outcomes of the ruling: overturn, uphold, or weaken Roe v.\ Wade by giving states more independent power to restrict abortions. The original paper used many experts’ opinions from books and law journals to determine how each justice will weight each criterion and how each sub-criterion will affect the alternative selected.

\begin{figure}
\begin{subfigure}[b]{\textwidth}
    \centering
    \includegraphics[scale=0.42]{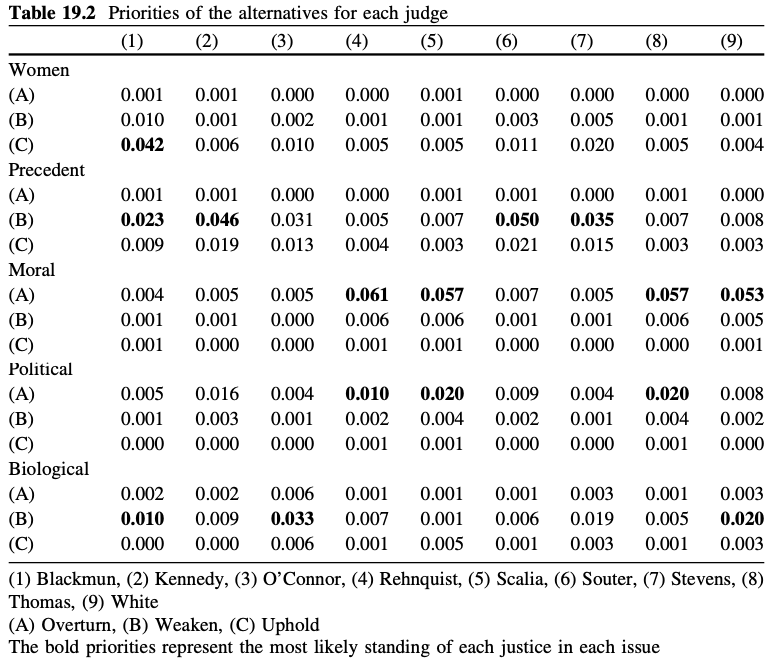}
\caption{Trace of synthesis of priorities. Figure taken from~\cite{Saaty2012}.}
    \label{fig:synthesis_court}
\end{subfigure}
\\~~\\
\begin{subfigure}[b]{\textwidth}
\begin{center} 
\includegraphics[scale=1.0]{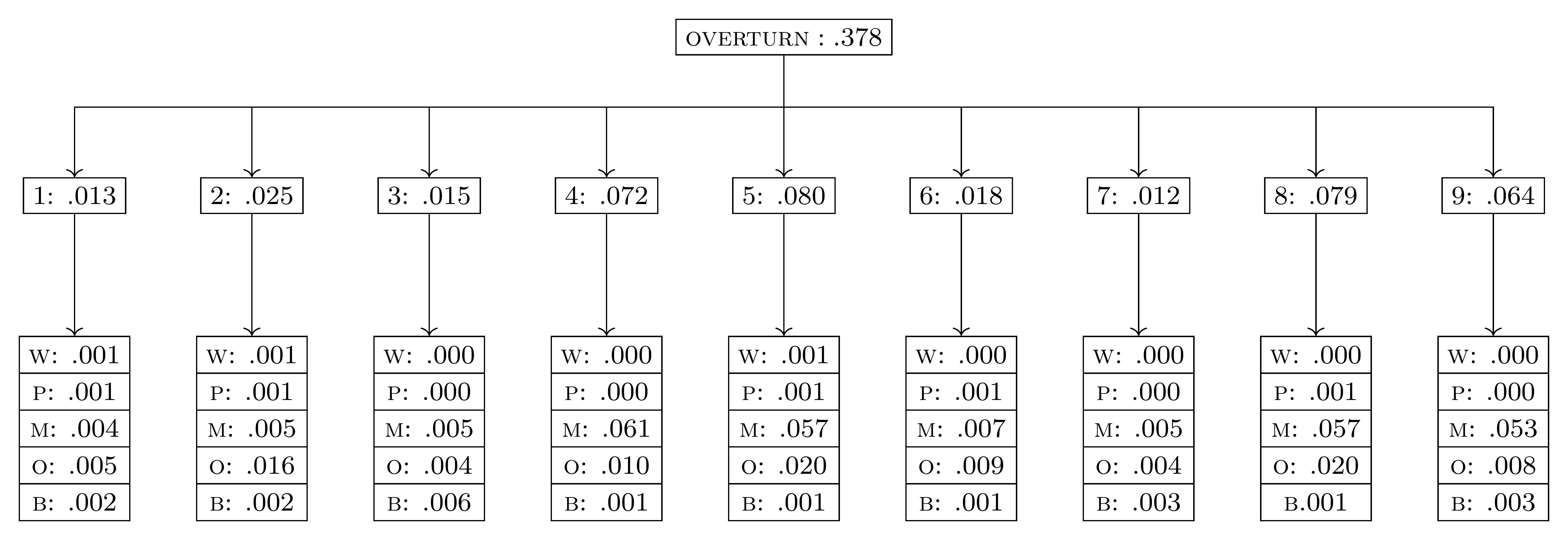}
 \end{center}

\caption{VD tree for the \overturn\ alternative.}
    \label{fig:vdtreecourt}
\end{subfigure}
    \caption{Synthesis of priorities in the Roe v.\ Wade example}
    \label{fig:my_label}
\end{figure}

\begin{figure}[t]
\begin{center} 
\includegraphics[scale=1.0]{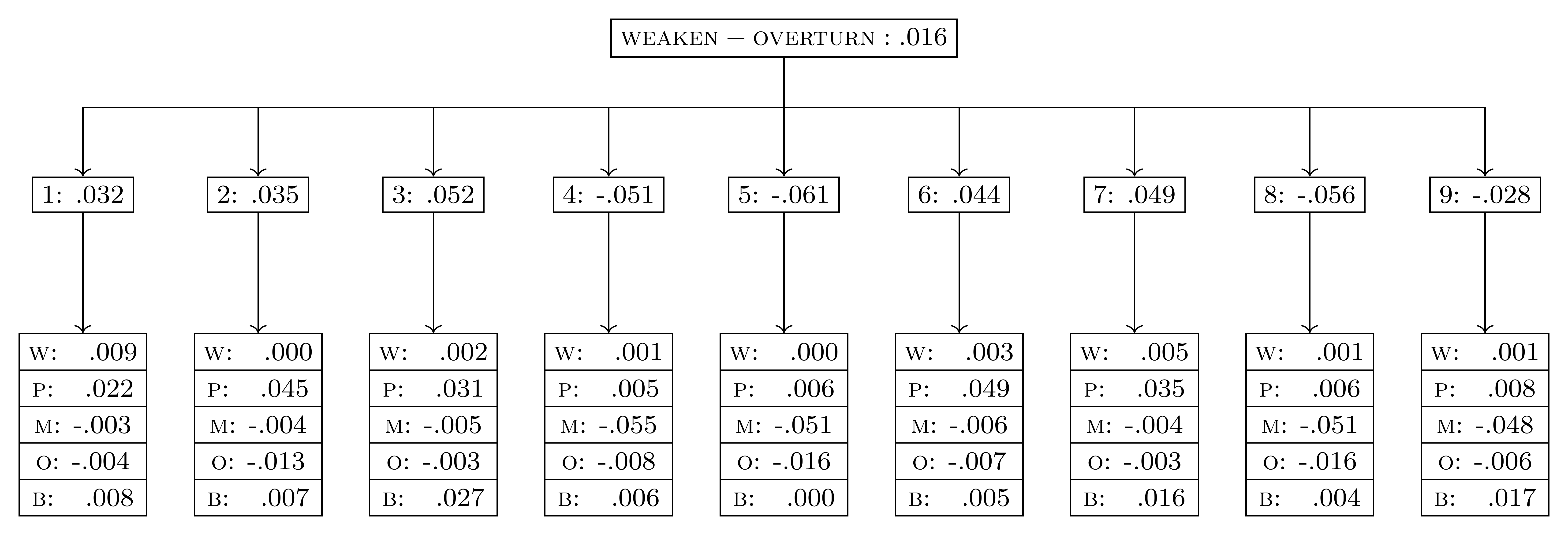}    
\end{center}

\caption{Hierarchical valuation difference between \overturn\ and \weaken\ in the Roe v.\ Wade example.}
    \label{fig:hvdtreecourt}
\end{figure}

The synthesis of priorities gives us priorities of $0.378$, $0.394$, and $0.228$ for overturn, weaken and uphold, respectively. The weaken rather than the overturn verdict is surprising, given the strong conservative leaning of the court. We can try to explain this surprising decision using our techniques. As a first step, we need to compute the hierarchical valuation difference. However, the original paper doesn't provide the matrix between levels 3 and 4 of the AHP model in Figure \ref{Court_Decision}. Due to lack of this matrix, we can't create the VD trees for various alternatives in the usual way. Interestingly, the paper provides a trace of the priority synthesis step as shown in Figure \ref{fig:synthesis_court}, which facilitates the computation of the VD trees for various alternatives, as shown for the \overturn\ alternative in Figure \ref{fig:vdtreecourt}.

Once we have the VD trees for the two alternatives, we can compute the hierarchical valuation difference between the two alternatives, as shown in Figure \ref{fig:hvdtreecourt}. The barrier for the decision shows that moral considerations and political leaning of all the 9 justices are for overturning Roe v.\ Wade rather than weakening it.
\begin{align*}
& \barr{\weaken,\overturn}=\\
 &~~\{ (1,\moral) \mapsto -.003, (2,\moral) \mapsto -.004, (3,\moral) \mapsto  -.005, (4, \moral) \mapsto -.055,(5, \moral) \mapsto -.051, (6, \moral) \mapsto -.006,\\
& ~~~~(7,\moral) \mapsto -.004, (8,\moral) \mapsto -.051, (9,\moral) \mapsto  -.048, (1, \political) \mapsto -.004,(2, \political) \mapsto -.013, (3, \political) \mapsto -.003,\\
&~~~~(4, \political) \mapsto -.008,(5, \political) \mapsto -.016, (6, \political) \mapsto -.007, (7, \political) \mapsto -.003,(8, \political) \mapsto -.016, (9, \political) \mapsto -.006 
\}
\end{align*}
However, the respect for precedent for all justices, ambiguity defining viability for all but justice 5 (Scalia) along with the consideration of the women issues by justices 1 and 7 (Blackmun and Stevens, respectively) lead to overcoming the barrier. This is a minimal explanation for why a weaken rather than an overturn verdict was reached. 
\begin{align*}
& \mdomi{\weaken-\overturn} = \\
& ~~\{ (1,\precedent) \mapsto .022, (2,\precedent) \mapsto .045, (3,\precedent) \mapsto  .031, (4, \precedent) \mapsto .005,(5, \precedent) \mapsto .006, (6, \precedent) \mapsto .049,\\
&~~~~ (7,\precedent) \mapsto .030, (8,\precedent) \mapsto .006, (9,\precedent) \mapsto .008, (1, \biological) \mapsto .008,(2, \biological) \mapsto .007, (3, \biological) \mapsto .027,\\
&~~~~(4, \biological) \mapsto .006,(6, \biological) \mapsto .005, (7, \biological) \mapsto .016,(8, \biological) \mapsto .004, (9, \biological) \mapsto .017,(1,\women) \mapsto .009,(7 ,\women)\mapsto.005
\}
\end{align*}
Since barrier as well as MDS contain a large number of components, focusing on specific levels of the hierarchy can simplify the explanation. An explanations which focuses on the sub-criteria used by justices is based on level 3 of the value difference. 
\begin{align*}
\xdiff3{\weaken-\overturn} = 
\{\women \mapsto .022,\precedent \mapsto .207,\moral \mapsto -.227 ,\political \mapsto -.076,\biological \mapsto .090\}    
\end{align*}
The corresponding barrier and MDS explanation are shown below.
\begin{align*}
& \xbarr3{\weaken-\overturn} = 
\{\moral \mapsto -.227 ,\political \mapsto -.076\} \\
& \xmdomi3{\weaken-\overturn} = 
\{\women \mapsto .022,\precedent \mapsto .207,\biological \mapsto .090\}  
\end{align*}
We observe that although moral consideration and political affiliation of the judges supports overturn of Roe v.\ Wade, their consideration for women's issues, precedent, and difficulty around defining viability outweigh this support, resulting in the less extreme verdict of weaken. It is interesting to note that a 7-2 conservative-leaning Supreme Court decided to just weaken Roe v.\ Wade in 1992 whereas a 6-3 conservative-leaning court overturned it in 2022. 

We can find out which judges were responsible for the weaken verdict by focusing on level 2. 
\begin{align*}
\xdiff2{\weaken-\overturn} = \{ & 1 \mapsto .032, 2 \mapsto .035,3 \mapsto .052 ,4 \mapsto -.051,5 \mapsto -.061, 6 \mapsto .044,\\
 & 7 \mapsto .049, 8 \mapsto -.056, 9 \mapsto -.028\}
\end{align*}
The barrier and MDS explanation tell us that justices 4, 5, 8, and 9 would most probably vote to overturn Roe v.\ Wade, however, the majority will prefer to weaken but uphold it. The final verdict \citep{PPCASEY} showed the same voting pattern as predicted by our explanation here. 
\begin{align*}
& \xbarr2{\weaken,\overturn} = 
\{4 \mapsto -.051,5 \mapsto -.061,8 \mapsto -.056, 9 \mapsto -.028\} \\
& \xmdomi2{\weaken,\overturn} = 
\{1 \mapsto .032, 2 \mapsto .035,3 \mapsto .052 , 6 \mapsto .044, 7 \mapsto .049\}
\end{align*}

\section{Evaluation}
\label{sec:eval}


To assess the effectiveness of MDS explanations for AHP decisions, we have performed a number of experiments to estimate the reduction in complexity that they can be expected to deliver. 
In the following we describe the setup and results of these experiments.

First, we have to establish criteria to measure the efficacy of explanations. Without any specific explanation, a user has to inspect all $n$ components of a value decomposition generated by the AHP process. The explanatory strength of an MDS comes from the fact that it can often reduce this number considerably to, say, $m$. The reduction can then be captured by defining the \emph{explanatory ratio} of an MDS as $m/n$.
The smaller the ratio, the fewer components users have to look at, relative to the original decision, thus making it easier to understand.
We can express the same idea more intuitively as a percentage size reduction achieved.
We thus define the \emph{MDS reduction} as
$R = (1-m/n)\times 100$, that is, an explanation ratio of 0.15 translates into a reduction by 85\%. 
%

\subsection{Efficacy of MDS Explanations}

Since there are no AHP benchmark data sets available, we have generated data for evaluating the efficacy of MDS explanations.
The examples reported in the literature indicate that AHP models rarely have more than 6 levels. Yet, each dimension can be wide: For example, an AHP model with 51 attributes in one dimension can be found in \citep{Liu2008ApplyingTA}. In general, it is common for an AHP to have about 10 attributes in one of the dimensions \citep{Pan2008FUZZYAA}. 

Based on these observations, we have randomly generated data for AHP examples having between 3 and 6 levels and computed the reduction for each case. A 3 level AHP is essentially a linear MCDM. 
The examples in the literature suggest limiting the number of components to 30 for models with 3 levels. For problems with 4, 5, or 6 levels, we limit the total number of components in the corresponding value decomposition of an AHP to 100, with intermediate dimensions having between 2 to 10 components each. 
We have used 20,000 random inputs for each scenario. For an AHP problem with a fixed number of levels, the inputs vary in two regards: (a) the number of attributes at each level, and (b) the values of the decision matrices. 
%

Another aspect that should be reflected in the test data is whether an explanation is necessary at all. For example, when the first alternative from an AHP process is better  than the runner-up in every regard, no explanation is necessary. In contrast, an explanation is most helpful in cases when the two alternatives are really close, that is, when the priority values of the alternatives are similar. 
To reflect this situation, we filter out those cases whose first two alternatives are not close. We call the relative difference between the priority values of two alternatives their \emph{decision margin} and consider scenarios in which the decision margin is bounded to 1\%, 5\%, 10\%, 20\%, and 30\%.

\definecolor{orange}{RGB}{253, 152, 16}
\definecolor{gray}  {RGB}{105, 105, 105}
\definecolor{blue}  {RGB}{16, 0, 251}
\definecolor{green} {RGB}{19, 116, 12}
\definecolor{purple}{RGB}{111, 0, 113}
\newcommand{\margin}[2]{\textcolor{#1}{\rule[0.5ex]{1em}{2pt}} $\leq #2\%$}
\newcommand{\tmargin}[2]{\textcolor{#1}{\rule[0.5ex]{1em}{2pt}} $#2$}


\begin{figure*}
\begin{subfigure}[b]{.45\textwidth}
    \includegraphics[scale = 0.23]{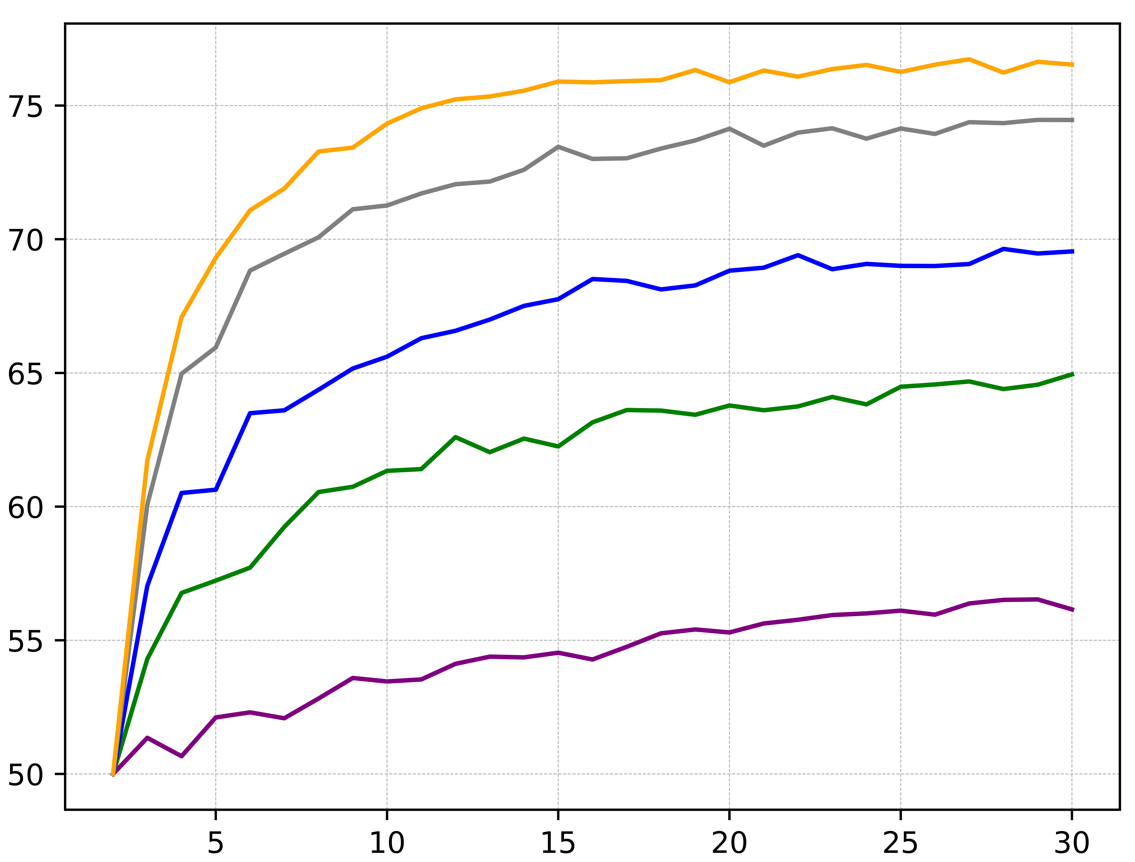}
    \caption{Number of levels: 3}
    \label{fig:Lev3}
\end{subfigure}
\begin{subfigure}[b]{.45\textwidth}
    \includegraphics[scale = 0.23]{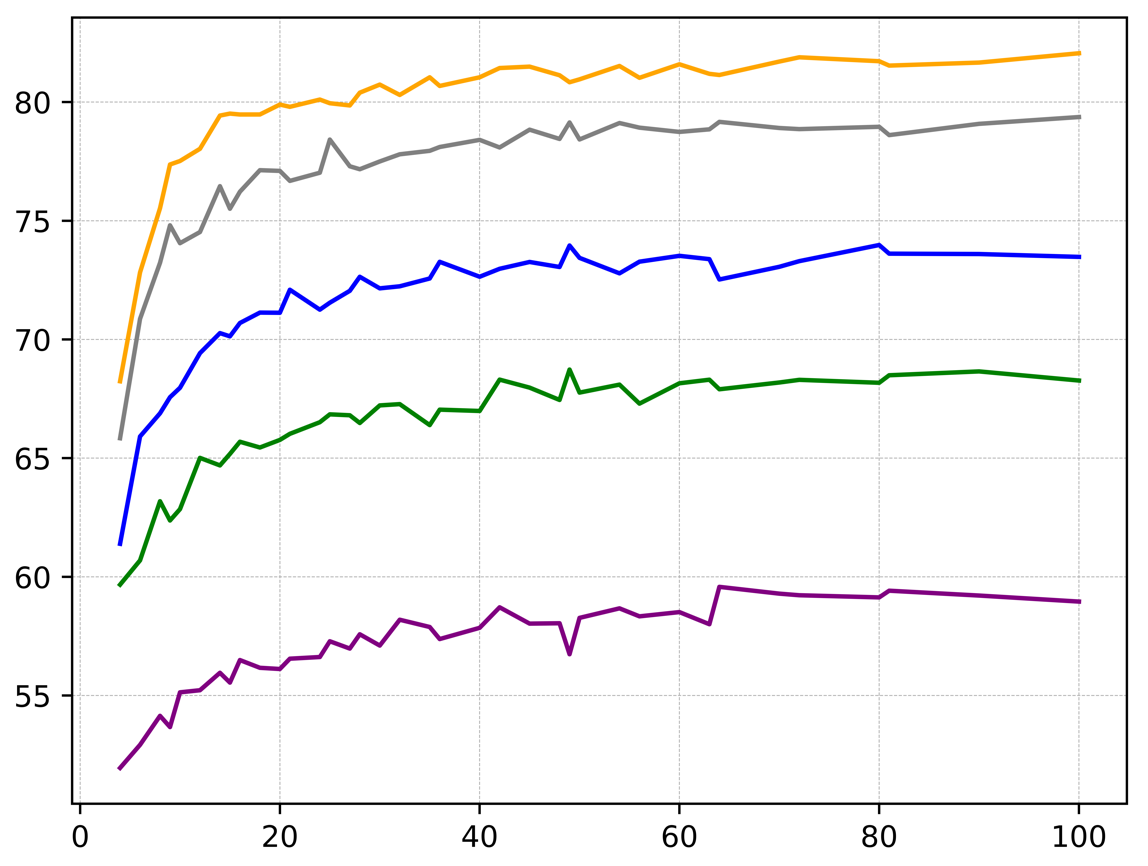}
    \caption{Number of levels: 4}
    \label{fig:Lev4}
\end{subfigure}
\newline\noindent\vspace{-1ex}
\begin{subfigure}[b]{.45\textwidth}
    \includegraphics[scale = 0.23]{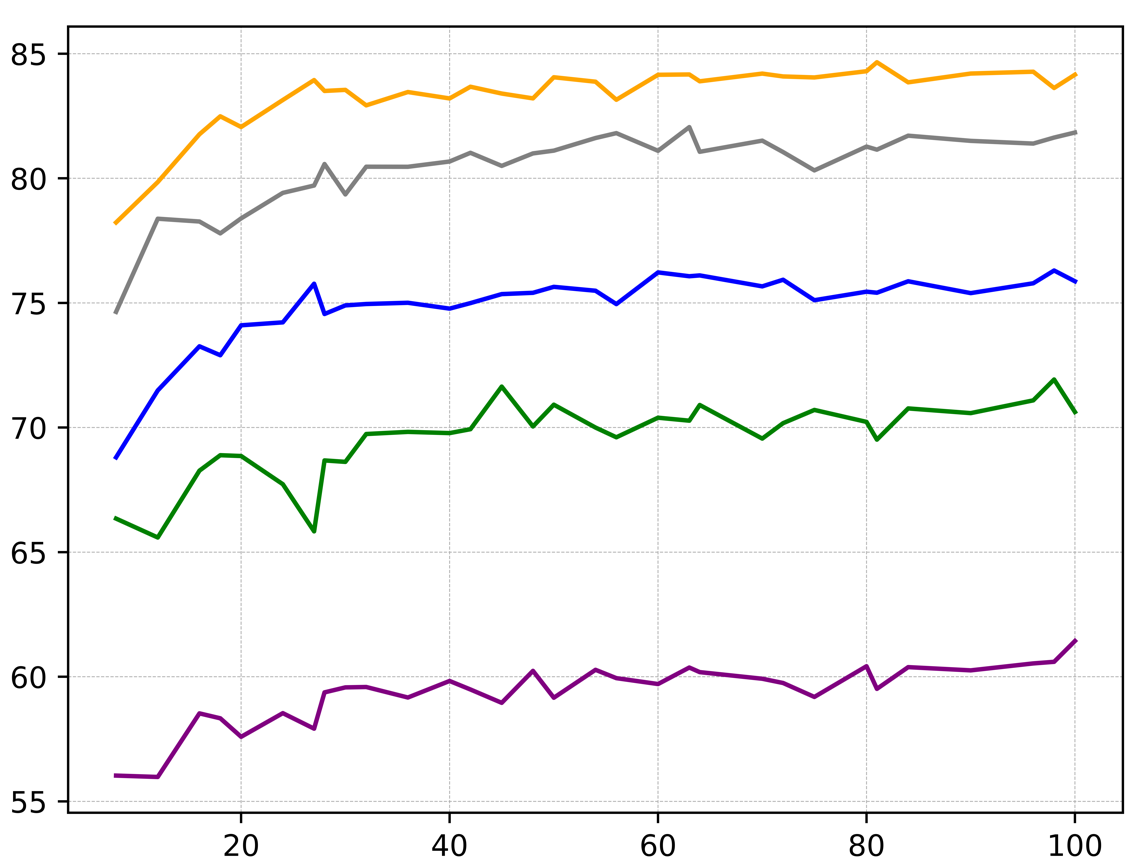}
    \caption{Number of levels: 5}
    \label{fig:Lev5}
\end{subfigure}
\begin{subfigure}[b]{.45\textwidth}
    \includegraphics[scale = 0.23]{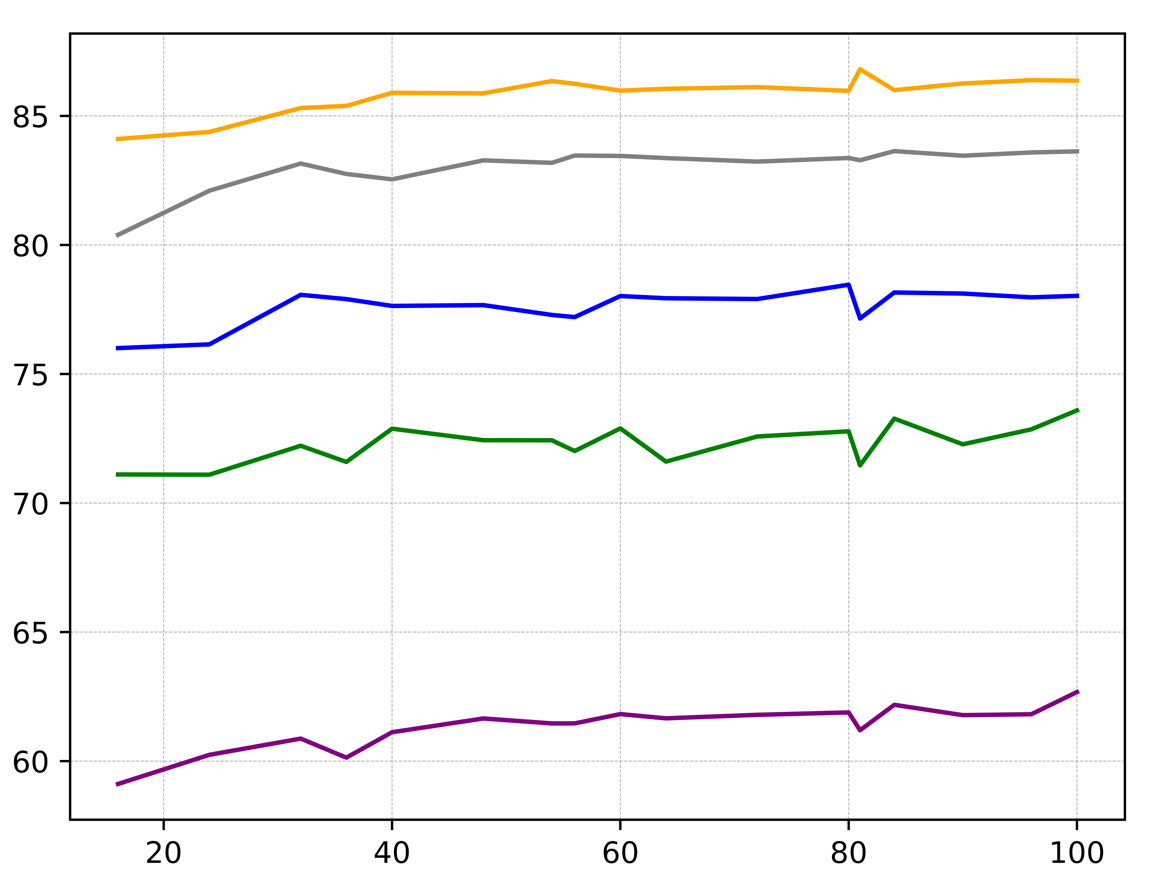}
    \caption{Number of levels: 6}
    \label{fig:Lev6}
\end{subfigure}
\caption{Average MDS reduction ($y$ axis) dependent on the number of components ($x$ axis) for AHP problems with different number of levels. 
Decision margins: \margin{orange}{30}, \margin{gray}{20}, \margin{blue}{10}, \margin{green}{5}, \margin{purple}{1}}
\label{fig:ReductionFactorGraph}
\end{figure*}


Figure \ref{fig:ReductionFactorGraph} shows how MDS reduction varies with the total number of components. We show graphs for AHPs with different number of levels containing plots for different decision margins.

The plots reveal some interesting trends. 
First, on average an MDS can prune the number of components by about 55-60\% even for a decision margin as low as 1\%. 
Second, the reduction decreases with smaller decision margins, which makes intuitive sense, since a greater value distance between alternatives provides more opportunities to explain the difference with fewer components.
But unfortunately, this also means that the efficacy of MDS explanation shrinks when they might be needed most.
Third, with an increasing number of levels, the curves ``move upward'', that is, for a given decision margin the reduction increases with the number of levels in the AHP problems.
In other words, MDS explanations scale well with the structural complexity of AHP problems.

\subsection{Efficacy of Single-Level Explanations}

\begin{figure*}
    \centering
    \includegraphics[scale = 0.18]{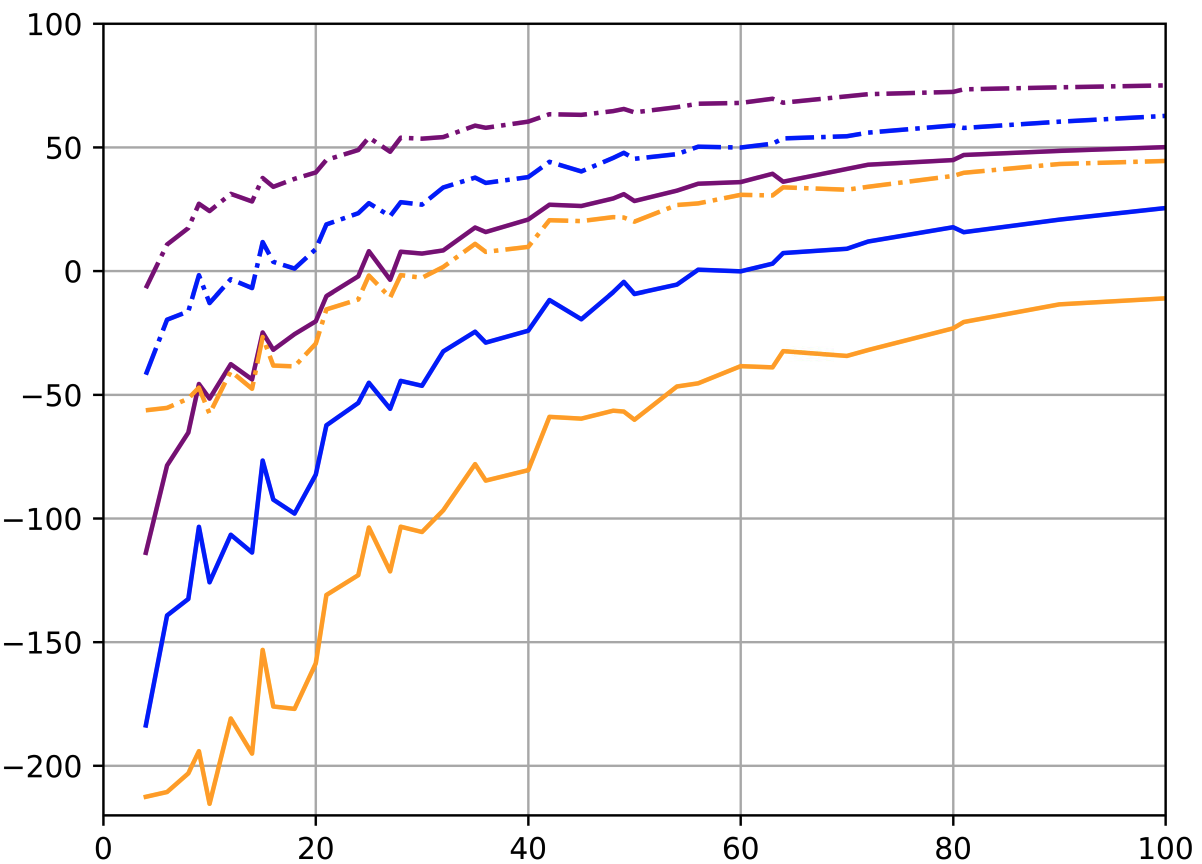}
    \caption{Size reduction of single-level explanations over MDS ($y$ axis) dependent on the number of components ($x$ axis) for AHP problems with 4 levels.
    Decision margins: \margin{orange}{30},  \margin{blue}{10}, \margin{purple}{1}
    (solid lines: worst case, dash-dot lines: average case).
}
    \label{fig:graph_coarse}
\end{figure*}



Consider the example from Section \ref{sec:rovevwade} where levels 2 and 3 consists of 9 and 5 components, respectively. Therefore, the total number of components in the value decomposition is 45 (= 9$\times$5), whereas a simplified explanation contains no more than the sum of the number of components in the two levels, that is, 14 (= 9 + 5).

We can measure the improvement of a single-level explanation over an MDS as the
size reduction given by $R= (1-s/m)\times 100$ where $m$ is the size of the MDS and $s$ is the size of the single-level explanation.
With this definition, single-level explanations promise an improvement in the Roe v.\ Wade example of at least 26\%.
The actual improvement was $(1-\frac{5+3}{19})\times 100 = 58$\%, which shows that the improvement can be significantly better than indicated by the worst-case estimate.

Figure \ref{fig:graph_coarse} shows the percentage improvement for simplified explanations over MDS. The $x$ axis shows the number of explanation components. For lack of space, we present the data only for AHP problems with 4 levels. However, the same trends can be observed for AHP problems with 5 and 6 levels. 

We can observe that for this worst-case consideration, savings can be obtained only for models with a significant number of components. But in general, the situation will be much better, since on average a single-level explanation will contain only half the number of components (because the other half will be used for the barrier).

\section{Related Work} 
\label{sec:RelatedWork}

Sensitivity analysis \citep{Triantaphyllou97asensitivity} is the tool of choice employed by decision makers to comprehend the results of various MCDM methods, including AHP. Sensitivity analysis is usually the only explanation mechanism available to a decision maker. 
Despite being useful, a potential limitation of sensitivity analysis is that it can only analyze the impact of one attribute at a time, keeping other attributes values constant. Thus, sensitivity analysis produces a number of localized explanations.
In comparison, our value-decomposition explanation method is global, and an MDS explanation takes into consideration the combined impact of various attributes in the decision, leading to generally more accurate and comprehensive explanations. 
On the other hand, MDS explanations are larger than the variation of one attribute, but the size of MDS explanations can be effectively reduced by employing single-level explanations.

The topic of explanations in general has been explored in a number of different areas.
While the origins of research into the nature of explanations can be traced back to philosophy \citep{Hempel65,Achinstein83,Ruben90}, the need for explaining computation has recently received a lot of attention, specifically in the area of AI \citep{Miller2019,Abadi2018}.



The notion of a value decomposition was introduced in \citep{EK21jfp} as a structure for explaining the results of dynamic programming algorithms. Value decomposition is generated as a domain-specific structure there, using the fact that dynamic programming algorithms can be viewed as instances of a mathematical semiring structure. This is similar to the current work, where the value decomposition is a domain-specific structure generated from the computations of the MCDM problems. Another point of similarity is that both generate contrastive explanations and thus also require two program results.
%
In \citep{EK21gpce} we describe a domain-specific language, which is based on the theory developed in this paper and allows users to specify MCDM problems, synthesize priorities for various alternatives, and generate MDS explanations. That work is primarily concerned with questions of language design and how to represent MCDM problems and explanations in support of computational transformations.

Similar to the current approach, another domain-specific structure created explicitly for explanations are provenance traces~\citep{acar2012}. A provenance trace consists of meta-information about the origin, history, or derivation of an object which is used in establishing trust and providing security in computer systems, particularly on the web.
Like value decompositions, provenance traces are a domain-specific explanation structure that works only in certain situations.

\section{Conclusions}
\label{sec:Conclusions}

We have demonstrated an effective method for explaining the results of MCDM methods.
Our approach of using minimal dominating sets is general enough to work well for flat and hierarchical models. Through the concept of single-level explanations, users have the option to additionally get simplified explanations.
As with explanations for algorithmic systems in general, the ability for generating concise explanations can contribute to the acceptance of results and adds transparency to computational systems. 


\section*{Acknowledgement}{This work is partially supported by the National Science Foundation under the grants CCF-1717300 and CCF-2114642.}


\bibliographystyle{apa}
\bibliography{BibFiles/References,BibFiles/explain,BibFiles/se,BibFiles/me,BibFiles/MADMReferences,BibFiles/explan_gen}  

\end{document}